\documentclass[conference]{IEEEtran}
\usepackage{times}
\usepackage{xcolor}

\usepackage[numbers]{natbib}
\usepackage{multicol}
\usepackage[bookmarks=true]{hyperref}
\usepackage{graphicx}
\usepackage{booktabs}
\usepackage{colortbl}
\usepackage{amsfonts}
\usepackage{amsmath,bm}
\usepackage{multirow}
\usepackage{mathtools}

\usepackage{lipsum}

\newcommand\blfootnote[1]{%
  \begingroup
  \renewcommand\thefootnote{}\footnote{#1}%
  \addtocounter{footnote}{-1}%
  \endgroup
}

\newcommand{\dd}[2]{\frac{\partial #1}{\partial #2}}
\newcommand{\ddline}[2]{\partial #1 / \partial #2}
\newcommand{\Loss}{\mathcal{L}}
\newcommand{\eg}{\textit{e.g.},~}
\newcommand{\ie}{\textit{i.e.},~}
\newcommand{\xw}{{\bf x}}
\newcommand{\nw}{{\bf n}}
\newcommand{\tw}{{\bf t}}
\newcommand{\change}[1]{{#1}}

\hypersetup{
    colorlinks=true,
    linkcolor=blue,
    filecolor=magenta,      
    urlcolor=magenta,
    pdftitle={An End-to-End Differentiable Framework for Contact-Aware Robot Design},
    pdfpagemode=FullScreen,
    }
    
\begin{document}

\title{An End-to-End Differentiable Framework for Contact-Aware Robot Design}






\author{\authorblockN{Jie Xu\authorrefmark{2},
Tao Chen\authorrefmark{2},
Lara Zlokapa\authorrefmark{2},
Michael Foshey\authorrefmark{2},
Wojciech Matusik\authorrefmark{2},
Shinjiro Sueda\authorrefmark{3} and
Pulkit Agrawal\authorrefmark{2}}
\authorblockA{\authorrefmark{2}Massachusetts Institute of Technology}
\authorblockA{\authorrefmark{3}Texas A\&M University}
\authorblockA{\href{http://diffhand.csail.mit.edu}{http://diffhand.csail.mit.edu}}}

\maketitle


\begin{abstract}
The current dominant paradigm for robotic manipulation involves two separate stages: manipulator design and control. Because the robot's morphology and how it can be controlled  are intimately linked, joint optimization of design and control can significantly improve performance. Existing methods for co-optimization are limited and fail to explore a rich space of designs. The primary reason is the trade-off between the complexity of designs that is necessary for contact-rich tasks against the practical constraints of manufacturing, optimization, contact handling, etc. We overcome several of these challenges by building an end-to-end differentiable framework for contact-aware robot design. The two key components of this framework are: a novel deformation-based parameterization that allows for the design of articulated rigid robots with arbitrary, complex geometry, and a differentiable rigid body simulator that can handle contact-rich scenarios and computes analytical gradients for a full spectrum of kinematic and dynamic parameters. On multiple manipulation tasks, our framework outperforms existing methods that either only optimize for control or for design using alternate representations or co-optimize using gradient-free methods.
\blfootnote{The video, appendix, supplementary document for simulation, and the code can be found in the project website: \href{http://diffhand.csail.mit.edu}{http://diffhand.csail.mit.edu}}
\end{abstract}

\IEEEpeerreviewmaketitle

\section{Introduction}
The design, control, and construction of manipulators is the cornerstone of robotics.
Today this process is manual and time-consuming as concurrent design of many different components is required. For example, hardware components and control algorithms are typically constructed sequentially making the integration of different modules difficult which necessitates many design iterations. Ensuring that the designed manipulator meets the desired specifications is challenging since there is a complex interplay between the robot design, manufacturing constraints, and the control algorithm. 

Due to the long iteration cycle, in practice, roboticists either (i) explore a rich design space, but make use of simple control algorithms \cite{deimel2016novel, rus2015design}
or (ii) develop complex algorithms to control existing robots \cite{akkaya2019solving, andrychowicz2020learning, nagabandi2020deep}.
The end result is a sub-optimal system for the given task. Co-optimizing both the design and the control scheme can significantly improve the performance of today's robotic systems. One challenge in co-optimization is the substantial increase in the number of parameters to be optimized. The other and arguably more significant challenge is in defining a representation of the robot design that is amenable to optimization. A good representation should: (a) result in designs that can be manufactured; (b) enable design of articulated robots with complex geometric shapes; and (c) allow the use of powerful optimization methods such as gradient descent. 

\begin{figure}[t!]
    \centering
    \includegraphics[width=\linewidth]{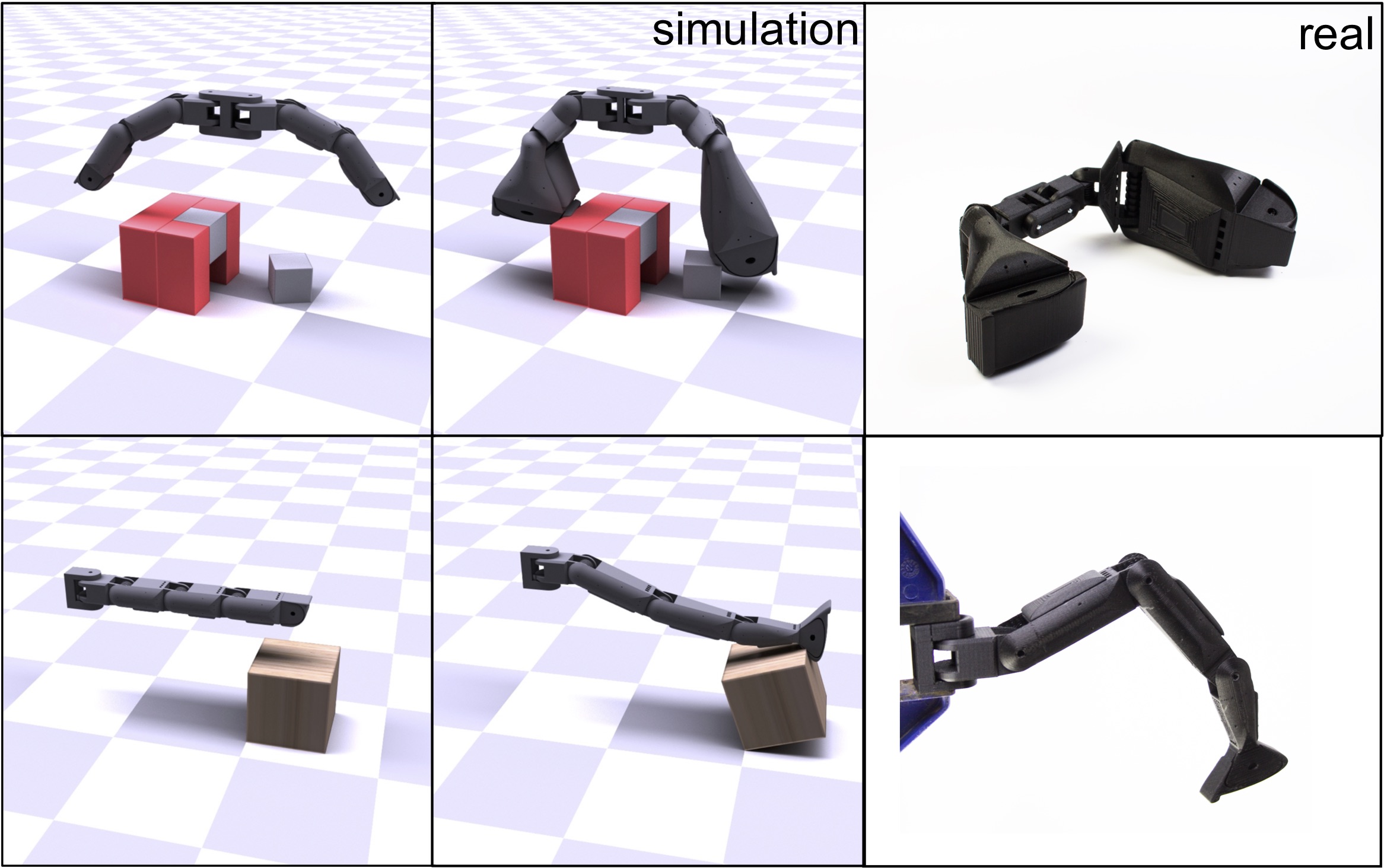}
    \caption{\textit{Left column}: only optimizing the control algorithm using a nominal robot design fails to complete the task; \textit{Middle}: co-optimization of morphology and control results in success; \textit{Right:} pictures of 3D-printed manipulators. Our method outputs designs that are easy to print and assemble.}
    \label{fig:intro}
    \vspace{-2mm}
\end{figure}

Several recent works have studied the design representation problem \cite{gupta2021embodied, ha2020fit2form, luck2020data,schaff2019jointly, zhao2020robogrammar}. One strategy is to represent the design as a graph, where each edge denotes connection between two components of the robot~\cite{gupta2021embodied, zhao2020robogrammar}. However, just defining the topology is insufficient: it is also necessary to parameterize the shape of each component. While there are many ways to represent the shape of a single object, the shape representation of an articulated system is a challenge. It is because the shape representation must span a rich space of geometric designs while simultaneously satisfying connectivity constraints between components and the manufacturing limitations. A popular strategy is to model each component by a simple primitive shape (\ie cylinder, cuboids, etc.)~\cite{gupta2021embodied, hazard2018automated, luck2020data, pan2020emergent, schaff2019jointly}. 
However, while easy to optimize, these primitives are often an over-simplification of the desired shape and are insufficient to model complex gripper designs as shown in Figure \ref{fig:intro}.

An alternative to basic shape primitives is the CAD parameterization. However, this approach has several downsides: generating models from CAD parameterization is slow \cite{schulz2017interactive}, and updates in CAD model parameters often results in failures such as the model being disconnected or even failure in generating the model.  
Other options for rich shape representations are voxel grid, point cloud, signed distance functions~\cite{ha2020fit2form}, etc. Unfortunately, these are not suitable for articulated robot design because it is hard to impose connectivity constraints between individual components. A drastic alternative to overcoming the connectivity problem without compromising rich shape parameterization is to directly optimize the shape of the entire robot instead of individual components. But this does not solve the problem either because now identifying individual joints and components, which is necessary for simulation and manufacturing, becomes a problem. 
Another consideration worth discussing is that for learning to control one must first import the shape into the simulator. While CAD designs can be easily imported they do not provide analytical gradients. Other representations discussed above must be first converted to a mesh, a step that is non-differentiable. Without analytical gradients one must rely on data inefficient gradient free methods for solving the shape optimization problem.

{\change{In this paper, we study the effective parameterization for the continuous shape morphology assuming a fixed robot topology.}} 
We propose a general morphology representation for articulated robot designs based on cage-based deformation (CBD) models. Cage-based deformations \cite{jacobson2011bounded, 10.1145/1186822.1073229} have been widely used in computer graphics to deform a mesh through a few number of cage ``handles'' {\change{(\ie cage vertices)}} in real-time while preserving local geometric features{\change{, as shown in Figures \ref{fig:overview} and \ref{fig:articulated_CBD}}}. Instead of specifying a large number of optimization parameters for modelling complex shapes, CBD maintains an \textit{expressive} shape design space with only a few parameters.
Furthermore, as we will describe in Section \ref{sec:parameterization} the cage representation allows imposition of a rich set of manufacturing and connectivity constraints. 
Most importantly, CBD is not tied to a specific shape representation and can be easily used with different representations such as meshes, point clouds etc. It is computationally inexpensive for inference, flexible (\ie user can easily control the degrees of freedom describing the shape by changing the number of cage handles), and differentiable.

To exploit the differentiability of the proposed deformation-based parameterization, we develop a differentiable articulated rigid body simulator for \textit{contact-rich tasks},
that makes three key improvements over prior work. First, we generalize the frictional contact model proposed by \citet{geilinger2020add} that only supports contact between a single dynamic body (e.g., the manipulator) and a static object (e.g., wall) to support contact between multiple dynamic bodies, a key requirement for object manipulation. 
Second, we propose a modified contact damping force that is free of discontinuities. Third, we derive the analytical gradients for a full spectrum of simulation parameters shown in Table \ref{tab:sim_params}, including the positions of contact points.

\begin{table*}[t!]
\caption{Summary of different Morphology Parameterization Methods}
\label{tab:compare_methods}
\newcommand{\bad}{\cellcolor{red!12}}
\newcommand{\ok}{\cellcolor{yellow!20}}
\newcommand{\good}{\cellcolor{green!12}}
\newcommand{\non}{\cellcolor{white!20}}
%
\newcommand{\checked}{\cellcolor{green}}
\vspace{-1em}
\scalebox{0.95}{
\begin{tabular}{c|cccccc}
\toprule
\non \textbf{Parameterization Method}&\non\textbf{Mesh Inference}&\non\textbf{Complex Shape}&\non\textbf{Differentiability}&\non\textbf{Dimension}&\non\textbf{Feature-preserving}&\non\textbf{Articulated Design}\\
\midrule
\midrule
\non Primitive Shapes \cite{ha2017joint, hazard2018automated, pan2020emergent, spielberg2017functional, wampler2009optimal} & \good Fast & \bad No & \good Support & \good Low & \bad No  & \good Support \\
\non CAD Parameterization \cite{10.1145/3355089.3356576, schulz2017interactive} & \bad Slow & \good Support & \bad No & \good Controllable & \good Yes & \bad No \\
\non TSDF \cite{ha2020fit2form} & \bad Slow & \good Support & \bad No & \bad High & \bad No  & \bad No\\
\non Mesh-based & \good Fast & \good Support & \good Support & \bad High & \bad No  & \bad No\\
\non Deformation-based (ours) & \good Fast & \good Support & \good Support & \good Controllable & \good Yes  & \good Support \\
\bottomrule
\end{tabular}
}
\vspace{-1em}
\end{table*}

The combination of the proposed deformation-based parameterization and the differentiable simulator allows us to build an end-to-end differentiable framework (Figure \ref{fig:overview}) for co-optimizing robot morphology and control for contact-rich manipulation tasks using analytical gradients. 
We test our framework on multiple manipulation problems, some of which are shown in Figure \ref{fig:intro}. The experiments show that our deformation-based parameterization provides us an expressive design space and the optimized designs can be easily manufactured (Figure~\ref{fig:intro} right). Furthermore, due to the key feature of end-to-end differentiability, the proposed method outperforms several state-of-the-art gradient-free approaches and model-free reinforcement learning methods at jointly optimizing the \textit{control scheme} and the \textit{robot morphology}.

\section{Related Work}
\subsection{Differentiable Physics-Based Simulation}
Physics-based simulation has been widely used for various robotic applications \cite{brockman2016openai, coumans2016pybullet, todorov2012mujoco}. Among them, differentiable physics-based simulators have gained increasing popularity recently since their differentiability facilitates efficient gradient-based optimization for robotic control. 
Due to their inherent differentiability, neural network based simulators have also been proposed to approximate physics \cite{battaglia2016interaction, chang2016compositional, li2018learning, mrowca2018flexible}. However, these works sacrifice generality and accuracy of physics for differentiability of the neural network.
Differentiable simulators have been developed for rigid bodies \cite{amos2019optnet, de2018end, degrave2019differentiable, geilinger2020add, heiden2019interactive, Wang2019}, soft bodies \cite{du2021diffpd, hahn2019real2sim, hu2019chainqueen, hu2020difftaichi, huang2021plasticinelab, jatavallabhula2021gradsim}, and cloth \cite{NEURIPS2019_28f0b864, qiao2020scalable}. Our simulator lies in the rigid body category, with some key modifications and improvements that enable contact-rich tasks through differentiability of frictional contact between multiple dynamic bodies. 
Several approaches have been proposed for making the frictional contact response differentiable such as differentiation of the coefficient matrices and vectors of the linear complementarity problem arising from collisions \cite{amos2019optnet, de2018end} or those that use impulse-based velocity stepping methods \cite{degrave2019differentiable}. However, the discontinuity stemming from such approaches can cause difficulties for contact-rich tasks such as ours. 
Recently, \citet{geilinger2020add} proposed a differentiable penalty-based frictional contact model. However, their contact model was only demonstrated to work for contact between the robot and a stationary surface (\eg ground and walls){\change{, and the simulation uses stiff springs to approximate articulation.}} 
We extend the frictional contact model to support a more general inter-object contact and develop a differentiable simulator based on the reduced coordinate formulation of RedMax \cite{Wang2019}, which allows a more compact representation {\change{and a more accurate articulated dynamics}}. More importantly, all the previous differentiable rigid body simulators only have gradient information for control variables and a small set of material parameters. In contrast, we derive the analytical gradients for a full spectrum of parameters, including the kinematic and dynamic parameters listed in Table \ref{tab:sim_params}. 

\subsection{Computational Robot Co-Design and Morphology Parameterization}

Co-design of robots typically involves optimizing geometry, mass properties, and control parameters. For articulated robot co-design, one must consider extra variables for kinematics and dynamics relationships among links, such as joint and body translations/orientations. Most existing works on robot co-design use simple primitive shapes to approximate the geometry of each robot link with gradient-based optimizations \cite{ha2017joint, hazard2018automated}, gradient-free approaches \cite{8202294, digumarti2014concurrent, meixner2019automated, pan2020emergent, wampler2009optimal}, or reinforcement learning (RL) based methods \cite{chen2020hardware, luck2020data, schaff2019jointly}. However, such over-simplified morphology parameterization based on primitive shapes precludes the possibility of generating complex geometric morphology for the robot. It is inadequate especially when the geometry does affect the task dynamics with a rich amount of contact, such as in in-hand manipulation tasks. In contrast, our deformation-based parameterization works for a wide range of complex shapes while preserving the geometric features, narrowing down the robot morphology representation gap between the simulation environment and real fabricated scenarios. CAD-based parameterization can also support natural organic shapes, but it typically suffers from its slow inference speed and non-differentiability.  \citet{schulz2017interactive} proposed an interactive system for CAD models with an expensive precomputation cost. \citet{10.1145/3355089.3356576} developed a differentiable parameterization of CAD models for FEM simulation, but the differentiability is not preserved in our rigid, multi-body simulation setting. Furthermore, CAD-based parameterization requires extra expertise to select and constrain each parameter in order to preserve model manufacturability throughout the optimization. It is also non-trivial to support the connectivity constraints required in an articulated robot structure. Truncated Signed Distance Functions (TSDF) and mesh-based parameterization are used by \citet{ha2020fit2form} to optimize the shape of a free-form gripper. These methods only work for a single body system, introduce a large number of parameters, and usually do not result in natural organic shapes. Compared to these, our deformation-based parameterization allows us to have a constrained but expressive design space for natural shapes and seamlessly works for articulated robot designs.
We summarize the comparison among different morphology parameterization methods in Table \ref{tab:compare_methods}.

\section{Method}
\begin{figure*}[t!]
    \centering
    \includegraphics[width=\linewidth]{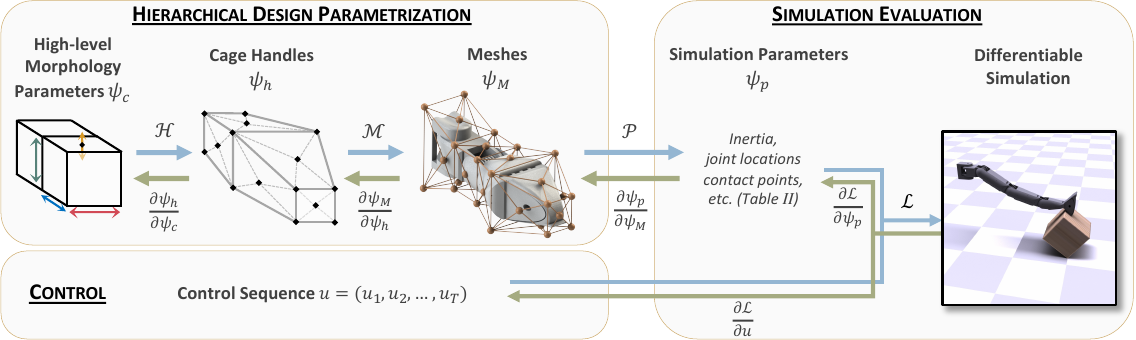}
    
    \caption{\textbf{End-to-end differentiable framework for morphology and control  co-optimization.} Blue arrows labeled as $\mathcal{H}, \mathcal{M}, \mathcal{P}$, and $\Loss$ are hierarchical functions that evaluate the loss function given the high-level morphology parameters, $\psi_c$ and controls, $u$. The corresponding green arrows are the derivatives.}
    \label{fig:overview}
\end{figure*}

{\change{We now describe our end-to-end differentiable framework for contact-aware robot design.
In Section \ref{sec:parameterization}, we describe our novel deformation-based design space for articulated robot morphology. A key insight of our morphology design space is the use of cage-based deformation, which allows us to morph the underlying mesh using a small number of cage ``handles''. More specifically, as shown in the top-left block of Figure \ref{fig:overview}, we use a two-level hierarchy to parameterize the shape. The high-level morphology parameters (shown in red, green, blue, and yellow arrows) controls the positions of the cage vertices (handles), and the cage handles in turn deform the underlying mesh.
In Section \ref{sec:simulation}, we describe our differentiable articulated rigid body dynamics simulation.
As shown in the right block of Figure \ref{fig:overview}, the simulation takes the deformed meshes and a control sequence as input, executes the forward steps, and computes the objective loss $\mathcal{L}$.
By combining the two key techniques above, in Section \ref{sec:framework}, we describe our end-to-end framework for robot design. Since each step is differentiable, the overall framework is differentiable, allowing us to use a gradient-based optimization method to search for the design parameters and the controls.}}

\subsection{Hierarchical Morphology Parameterization}
\label{sec:parameterization}
Morphology optimization relies on an effective morphology design space, which further depends on an effective morphology shape parameterization. In this section, we describe our approach to leverage the cage-based deformation as the morphology parameterization for articulated robots with \textit{complex} component shapes. {\change{As shown in the top-left block in Figure \ref{fig:overview}, we use a two-level hierarchy to parameterize the shape of the robot: the cage-based deformation $\mathcal{M}$ and high-level morphology parameterization $\mathcal{H}$.}}

\begin{figure}[t!]
    \centering
    \includegraphics[width=\linewidth]{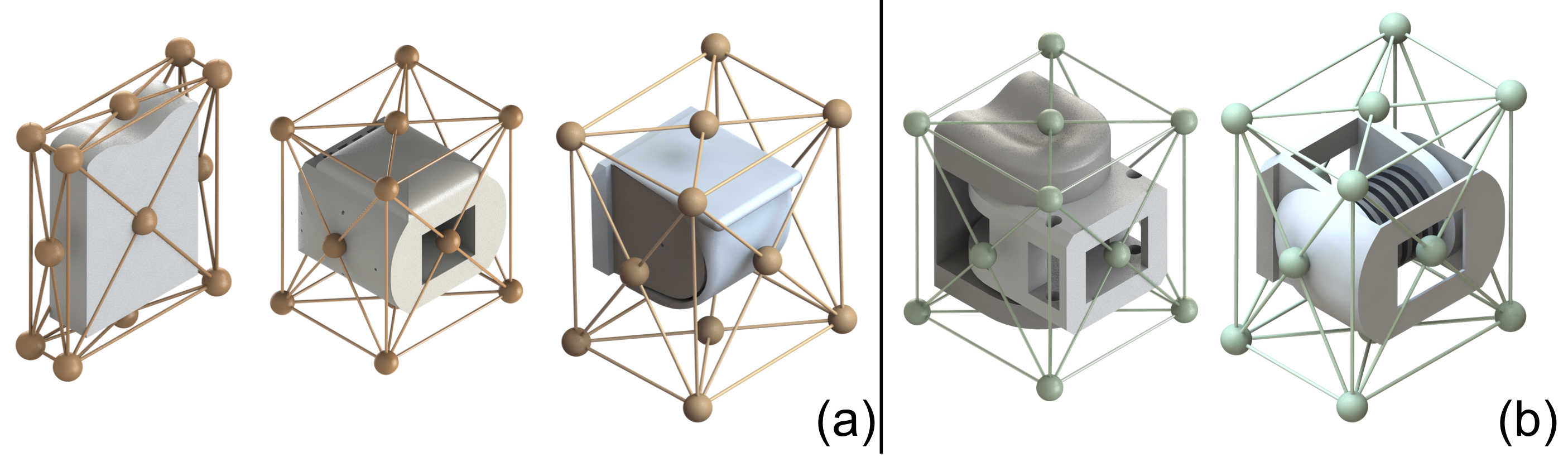}
    
    \caption{Our component database for our manipulators. From left to right: finger base, phalanx segment, finger tip, knuckle, and joint. Each component comes with its own deformation cage. (a) The components in the yellow cages can be deformed arbitrarily with the cage, whereas (b) the components in the green cages can only be expanded along the axis of rotation. }
    \label{fig:component_database}
    \vspace{-0.8em}
\end{figure}

\textbf{Cage-based Deformation} ($\mathcal{M}$)~
Cage-based deformation (CBD) is a classic geometry processing technique in computer graphics used to deform a high-resolution mesh in a real-time and feature-preserving manner. With a coarse, closed cage, CBD controls the enclosed space's deformation by moving the cage vertices, or \textit{cage handles} around. {\change{Let $\mathcal{C}$ denote the cage and $H$ denote the cage vertices (\ie handles) of $\mathcal{C}$ with $\tilde{\psi}_h$ being the positions of the handles in $H$ in the rest configuration, and let $\mathcal{S}$ be the space enclosed by cage $\mathcal{C}$. For any arbitrary point $\tilde{s} \in \mathcal{S}$, CBD computes a normalized barycentric coordinate $\bm{w} \in \mathbb{R}^{|H|}$ for the point, called \textit{deformation weights}, which satisfies:
\begin{equation}
    \tilde{s} = \sum_j^{|H|}\bm{w}_j\tilde{\psi}_h(j) \quad \text{and} \quad \sum_j^{|H|} \bm{w}_j = 1.
\end{equation}
These deformation weights then define a linear function to transfer the translation of handle vertices to the movement of the associated point at run time through:
\begin{align}
    s = \sum_j^{|H|}\bm{w}_j\psi_h(j),
\end{align}
where $\psi_h$ is the new positions of the cage handles, and $s$ is the new position of $\tilde{s}$ under deformation.}} The deformation weights are precomputed for each particular point in the space $\mathcal{S}$ and kept constant at run time. 
CBD methods preserve various features of the underlying, high-resolution mesh after deformation by carefully designing the weight construction algorithms. Among various CBD methods, we choose the mean value coordinates method \cite{10.1145/1186822.1073229} for its simplicity, stability, and capability to be extended to deform articulated structures. 

{\change{Inspired by its power, we apply cage-based deformation to parameterize the shape of each robot component by the positions of a set of cage handles around the shape mesh. In practice, the cage handles of each component can be defined by the users based on their demands. For our purpose in this work,}} we construct a component database for manipulator construction as shown in Figure \ref{fig:component_database}. Each component is represented by a mesh $M^i$ and is associated with a predefined cage $\mathcal{C}^i$ around it. We then use the mean value coordinates method to precompute the deformation weight matrix $D^i$ for each component mesh $M^i$, and reuse those weights afterwards. Let $V^i$ be the set of vertices of mesh $M^i$ and $H^i$ be the set of handles in cage $\mathcal{C}^i$. Then $D^i$ is a $|V^i|$-by-$|H^i|$ matrix storing the deformation weight for each vertex on the mesh with respect to each handle on the cage. 
The deformation weights precomputation by mean value coordinates method is cheap, with a computational cost $\mathcal{O}(|V^i|\cdot|H^i|)$. 
Such precomputation and the run-time linear combination enable a fast mesh inference given the new cage handle positions. 
Moreover, by controlling the high-resolution mesh's morphology through a coarse cage, we effectively reduce the morphology space into a relatively low-dimensional space for natural deformed shapes. This provides us a constrained yet expressive morphology design space. 
Furthermore, the dimension of the deformation is still fully controllable. By adding more handle points, one can deform the underlining mesh with more degrees of freedom, which makes it possible to find a good trade-off between low-dimensional morphology parameter space and fine-grained morphology deformation for different applications.
Most importantly, this linear combination gives us a fully differentiable function $\mathcal{M}$ mapping from handle positions $\psi_H$ to the positions of mesh vertices $\psi_M$. 

\begin{figure}[t!]
    \centering
    \includegraphics[width=\linewidth, height=5cm]{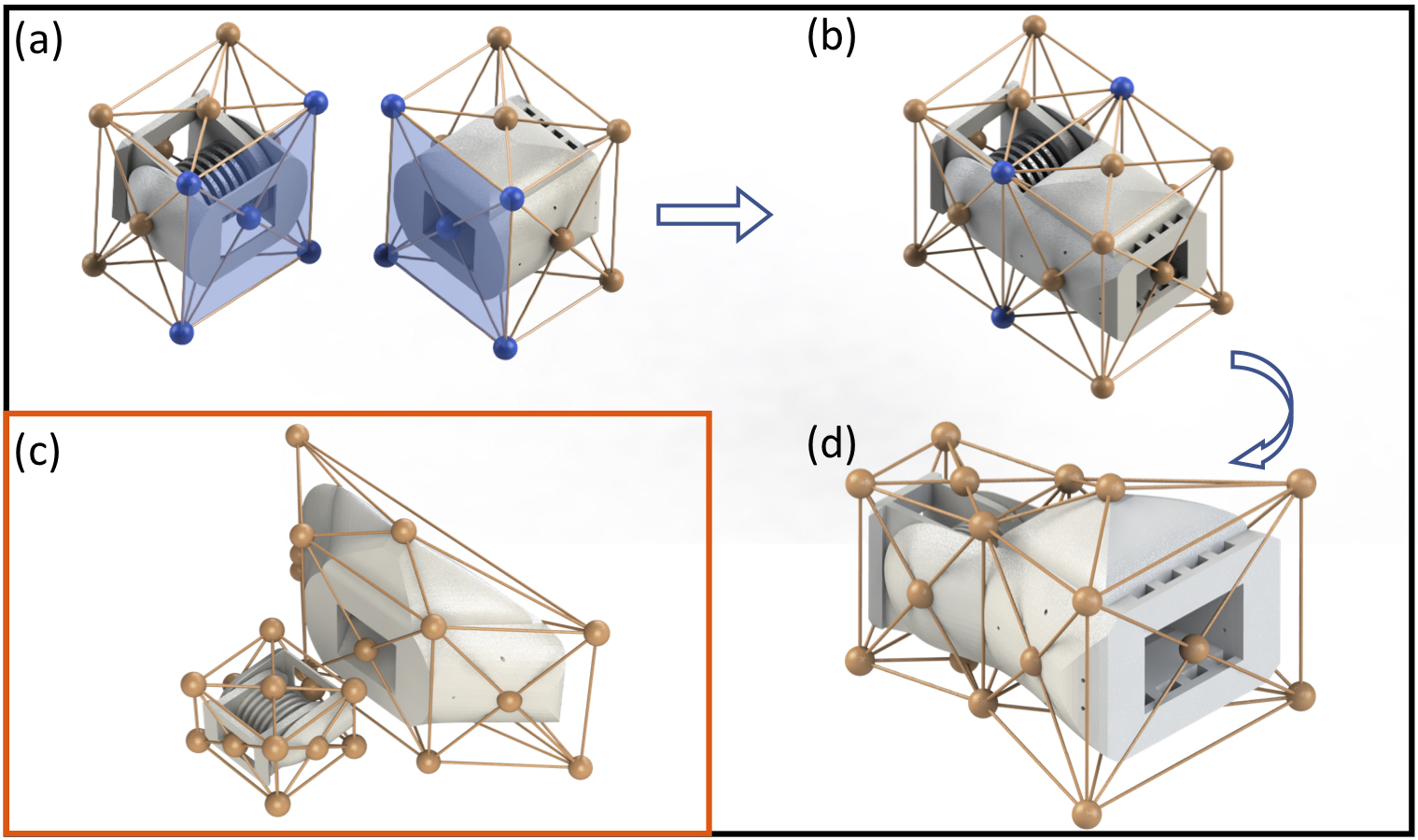}
    
    \caption{A \textit{joint} component and a \textit{body phalanx segment} component are shown in the figures. We parameterize the articulated components into lower-dimensional parameters $\psi_c$ by posing different deformation constraints on each component and merging their handle points on the connection surface (highlighted in blue in (a) and (b)). We can then freely explore the $\psi_c$ space to change the underlining articulated robot shape (d). The two components come apart from each other and become to be not manufacturable if they are deformed individually and arbitrarily by their associated cages (c).}
    \label{fig:articulated_CBD}
    \vspace{-1em}
\end{figure}

\textbf{High-level Morphology Parameterization} ($\mathcal{H}$)~
{\change{The CBD method described above works for any \textit{single} mesh with an arbitrary shape. 
However, parameterizing an articulated robot poses extra challenges, since}} independently manipulating the cage handles arbitrarily for different robot components will easily lead to a design that is not connected and not manufacturable, as shown in Figure \ref{fig:articulated_CBD}(c).
In order to handle proper articulation, we need to take two extra constraints into consideration: the \textit{fabrication constraint} and the \textit{connectivity constraint}. 

The \textit{fabrication constraint} requires us to have different deformation constraints for different components based on their manufacturing methods. For instance, the finger body part can be manufactured using a 3D printer, so it can undergo free-form deformation. In contrast, the joint component is usually composed of some commercially-sourced products (\eg screws, spring pins, etc.), which come in predefined, standard sizes. Thus, components such as joints can only be expanded in directions that maintain the geometric integrity of the model where it interfaces with these standard-sized parts.
We achieve this through further parameterizing the cage with a few extra parameters to control the allowed deformation for each component. 

\textit{Connectivity constraint} is a more challenging problem for articulated designs, requiring components to remain connected after deforming each underlying mesh. It is non-trivial for CAD-based and mesh-based parameterizations to satisfy this constraint since they are unable to track the connectedness of the interface surfaces between components while varying the parameters. 
We instead leverage a special property of mean value coordinates to resolve the connectivity issue. Let $\langle H_1, H_2, H_3\rangle$ be three handles of a triangle on the cage mesh $\mathcal{C}$, and $s$ be a point in the enclosed space. With mean value coordinates, if $s$ and $\langle H_1, H_2, H_3\rangle$ are coplanar, and $s$ is inside the triangle $\langle H_1, H_2, H_3\rangle$, then the deformation weights for $s$ have the following property:
\begin{equation}
    \label{eq:mvc_property}
    \bm{\omega}^s_i = \left\{\begin{array}{ll}
                    0, & \text{if } i \not\in  \{H_1, H_2, H_3\} \\
                    >0, & \text{if } i \in  \{H_1, H_2, H_3\}.
                    \end{array}
    \right.
\end{equation}
In other words, $s$ is fully controlled by the triangle that contains it.

With this property in mind, we construct the cage for each component so that, for each component's connection surface (\ie interface to a neighboring component), we always construct a cage that is coplanar with and fully contains the connection surface as shown by the blue faces in Figure \ref{fig:articulated_CBD}(a). We call the handles on the connection surface plane \textit{connection handles}. For two components that can be potentially connected, their connection handles are constructed to be identical, which ensures that the connection surfaces on two components share the same connection handle layout. While connecting two components, we merge their overlapped connection handles (Figure \ref{fig:articulated_CBD}(b)). Such merge operation is equivalent to adding a constraint on the handles on the connection surface so that their connection handles always have the same motion.
Due to the property of mean value coordinates in Eq.~\ref{eq:mvc_property}, this placement of connection handles ensures that the connectivity constraint is automatically satisfied.

By considering both the \textit{fabrication constraint} and the \textit{connectivity constraint}, we parameterize the whole cage 
via a small number of high-level morphology cage parameters $\psi_c$. {\change{This extra layer of parameters implicitly imposes constraints on all cage handle positions to move around in a unified fashion.}} Specifically, $\psi_c$ consists of: (a) the scale information of the cages (\eg length of each cage, and width/height of each connection cage, shown as red, blue, and green arrows in Figure \ref{fig:overview}); and (b) any other auxiliary cage points (\eg yellow vertical arrow in Figure \ref{fig:overview}). The scale parameters are component-dependent and are based on their fabrication constraint. For example, the joint cage can only be scaled along the joint axis direction. 
Through such cage parameterization, we can map the high-level cage parameters $\psi_c$ to the cage handle positions $\psi_h$ by $\psi_h = \mathcal{H}(\psi_h)$, and effectively construct a high-level morphology design space with only bound constraints. {\change{Note that}} this high-level cage parameterization is not a fixed choice; the parameterization can be modified based on the application and user's demand. We show a more free-form cage parameterzation in the experiment section to show such flexibility.

\change{Although different combinations of the components results in a large amount of different design topologies (\eg different number of fingers, different number of links, etc.) and thus different final deformation cages (\ie merged cages),
we construct a simple stochastic graph grammar, following the work by \citet{zhao2020robogrammar}, to generate different topologies and automatically construct the merged deformation cage and the high-level cage parameterization. Since we focus on the continuous morphology optimization for a fixed design topology in this work, we leave the details of the grammar in Appendix I.}

\subsection{Differentiable Articulated Rigid Body Simulation}
\label{sec:simulation}
From the deformation-based parameterization, we obtain the mesh vertices $\psi_M$. In order to simulate the constructed morphology, we further convert $\psi_M$ into the simulation parameters $\psi_p$ through an analytical function $\mathcal{P}$. As shown in Table \ref{tab:sim_params}, the simulation parameters $\psi_p$ include both kinematics- and dynamics-related parameters. Specifically, the kinematic parameters are the relative transformations of the joints with respect to their parent joints, $E_j$, and the relative transformations of the bodies with respect to their parent joint, $E_b$; and the dynamic parameters are the generalized inertia, $I$, contact point positions with respect to the bodies, $C_b$, and surface area for each contact point, $a$. For the generalized inertia, we use cuboids for ease of differentiability, but it is also possible to use mesh-based inertia. (Note, however, that the 3D printed parts may not necessarily match the mesh-based inertia, depending on the in-fill.) In order to acquire the contact points $C_b$ in the deformed mesh, we presample a uniformly distributed set of contact points on the surface of each mesh in the rest configuration. We then track the positions of these presampled contact points through the same cage-based deformation as the mesh vertices. Thus, the deformation-based parameterization provides us with differentiability not only for the mesh but also for the contact point positions. The approximate contact point area, $a$, is used to scale the magnitude of the frictional contact forces. To compute this parameter, we use the change in the total surface area of the cage before and after deformation.

\begin{table}[t]
\caption{List of simulation parameters $\psi_p$}
\centering
\label{tab:sim_params}
\vspace{-1em}
\begin{tabular}{cccc}
\toprule
\textbf{Type} & \textbf{Notation} & \textbf{Parameter Description} & \textbf{Dimension} \\ 
\midrule
\midrule
\multirow{2}{*}[-0.5ex]{Kinematics} & $E_j$ & Joint transformation & $\mathrm{SE}(3) \times n_b$ \\ 
\cmidrule{2-4}
&$E_b$ & Body transformation & $\mathrm{SE}(3) \times n_b$ \\
\midrule
\multirow{3}{*}[-1.2ex]{Dynamics} & $I$ & Generalized inertia & $n_\text{DOF}$ \\
\cmidrule{2-4}
&$C_b$ & Contact points on body & $3 \times n_c$ \\
\cmidrule{2-4}
&$a$ & Contact area & $n_c$\\
\bottomrule
\end{tabular}
\vspace{-1em}
\end{table}

{\change{Unlike the previous methods which typically compute the gradients of the simulation only with respect to control parameters $\partial\mathcal{L}/\partial u$ or a small set of material parameters, our simulator also provides the analytical gradients $\partial\mathcal{L}/\partial\psi_p$ for a full spectrum of simulation parameters described above. Such extension is non-trivial and is essential for allowing gradient-based morphology optimization. We provide detailed mathematical formulation of our differentiable simulation in the supplementary document associated with the released code, and briefly introduce the key ideas of our simulation below.}}

\textbf{Simulation Dynamics}~
The simulation parameters $\psi_p$ and the control sequence $u$, are the input to the differentiable articulated rigid body simulator. Our simulator uses reduced coordinates so that the equations of motion are expressed compactly using a minimal set of degrees of freedom \cite{Wang2019}. {\change{The dynamics}} equations are implicitly integrated in time with the BDF2 scheme, with SDIRK2 for the initial step \cite{Nishikawa2019}. We analytically derive all the derivatives required by these implicit time integration schemes, and we solve the resulting non-linear equations using Newton's Method with line search. 
To compute the simulation derivatives, $\ddline{\Loss}{\psi_p}$ and $\ddline{\Loss}{u}$, we use the adjoint sensitivity method \cite{geilinger2020add,McNamara2004}, which requires a forward pass and a backward pass of the simulation. During the forward pass, we store some auxiliary variables, such as the matrix factors of the final Newton iteration of each time step and the partial derivatives of the loss function. Then during the backward pass, we compute the final derivative with a block banded triangular solver using these auxiliary variables stored during the forward pass.

\textbf{Inter-Object Frictional Contact Model}~
For frictional contact, we extend the differentiable penalty-based approach of \citet{geilinger2020add}, adapted to our reduced coordinate approach. We make two critical changes to make their penalty-based approach work in our manipulation settings.

First, we add support for frictional contact between \textit{two dynamic} bodies, rather than between a single dynamic body and a stationary surface (\eg floor and walls). With our approach, as we change the cage parameters, the contact point positions change, and so to allow end-to-end differentiability, we require the gradients of the normal contact force, the dynamic friction force, and the static friction force with respect to these contact point positions as well as with respect to the generalized coordinates of the robot (\ie joint angles). To solve this, we use a signed distance field augmented with derivative information. This distance field is attached to the manipulated object, which we assume is rigid. Let the generalized coordinates of the robot and the manipulated object be $q_b$ and $q_o$, respectively, and the generalized velocity be $\dot{q}_b$ and $\dot{q}_o$, respectively. Then the world position and velocity of the contact point $C_b$ on the robot body are computed as $\xw(q_b, C_b)$ and $\dot\xw(q_b, \dot{q}_b, C_b)$. Using this world point, we query the signed distance function attached to the manipulated object, which gives us the following quantities:
\begin{equation*}
\begin{matrix*}[l]
    d, \, \dot{d} & \text{penetration distance and speed}\\
    \nw           & \text{contact normal}\\
    \dot{\tw}     & \text{tangential velocity}\\
\end{matrix*}
\end{equation*}
as well as the \textit{derivatives} of these quantities with respect to $q_b$, $q_o$, and $C_b$. The frictional contact forces are computed from these quantities and then scaled by the approximate contact area, $a$. The derivatives of the frictional contact forces with respect to $q_b$ and $q_o$ are used to step the simulation forward in time with the implicit integrator, and the derivatives with respect to $C_b$ and $a$ are used to calculate the sensitivities during the backward pass of the simulation.

Second, we modify the contact damping force so that it is continuous. The original formulation by \citet{geilinger2020add} has a discontinuity, which we found can cause convergence problems when there are collisions between two moving dynamic objects. In their formulation, the contact damping force is proportional to the penetration speed, $\dot{d} \nw$, which means that there will be a sudden change in the magnitude of this damping force at $d = 0$, since $\dot{d}$ is not necessarily zero when $d$ is zero. To fix this, we instead make the damping force be proportional to both the penetration distance and speed, $d \dot{d} \nw$, which ensures that the force remains continuous at the moment of collision. If needed, we can also use a sigmoid function around $d$ to make the damping force match the original formulation when $d$ becomes large.

\subsection{End-to-End Differentiable Co-Design Framework}
\label{sec:framework}

{\change{By combining the proposed deformation-based parameterization and the differentiable simulator, we build an end-to-end differentiable framework for robot co-design as shown in Figure \ref{fig:overview}.}}

{\change{Mathematically speaking, our co-design framework starts with a three-layer morphology parameterization $\mathcal{F} = \mathcal{P} \circ \mathcal{M} \circ \mathcal{H}: \mathbb{R}^m\rightarrow\mathbb{R}^{|H|\times3}\rightarrow\mathbb{R}^{|V|\times 3}\rightarrow \mathbb{R}^p$, where $m$ is the number of high-level morphology parameters, $|H|$ is the total number of cage handles, $|V|$ is the total number of mesh vertices, and $p$ is the number of low-level kinematic and dynamic parameters in the simulation. 
This hierachical parameterization converts high-level morphology parameters $\psi_c$ into low-level simulation parameters $\psi_p$, going through three analytically differentiable steps including the morphology parameterizations $\mathcal{H}$ and $\mathcal{M}$ in Section \ref{sec:parameterization}, and the simulation parameter computation $\mathcal{P}$ in Section \ref{sec:simulation}:
\begin{equation}
    \psi_p = \mathcal{P}(\psi_M),\; \psi_M = \mathcal{M}(\psi_h),\; \psi_h = \mathcal{H}(\psi_c).
\end{equation}
The differentiability of each step allows us to efficiently compute the derivatives from the simulation parameters $\psi_p$ all the way to the high-level morphology parameters $\psi_c$ through the chain rule:
\begin{equation}
    \dd{\psi_p}{\psi_c} = \dd{\psi_p}{\psi_M} \dd{\psi_M}{\psi_h} \dd{\psi_h}{\psi_c}.
\end{equation}
The framework then proceeds with the simulation described in Section \ref{sec:simulation} with the simulation parameters $\psi_p$ and a control sequence $u$ as input, and computes the task-specific objective loss $\mathcal{L}$. As our simulator is differentiable with respect to both the kinematic/dynamic parameters and control variables, we are able to compute the analytical derivatives $\dd{\Loss}{\psi_c} = \dd{\Loss}{\psi_p} \dd{\psi_p}{\psi_c}$ and $\dd{\Loss}{u}$ for the full framework efficiently.}}

\section{Experiments}
\label{sec:experiments}

\subsection{Implementation}
We implemented our differentiable rigid body simulation in C++ and the deformation-based design parameterization in Python. {\change{The two components of the code are connected through Python bindings.}} The control sequence input to the simulation consists of the torques applied to the joints at each simulation time step. In all the tasks shown in below, we use L-BFGS-B \cite{nocedal2006numerical} for co-optimization with our analytical derivatives, $\ddline{\Loss}{\psi_c}$ and $\ddline{\Loss}{u}$. 

\subsection{Quality of Morphology Design Space}
\begin{figure}[t!]
    \centering
    \includegraphics[width=\linewidth]{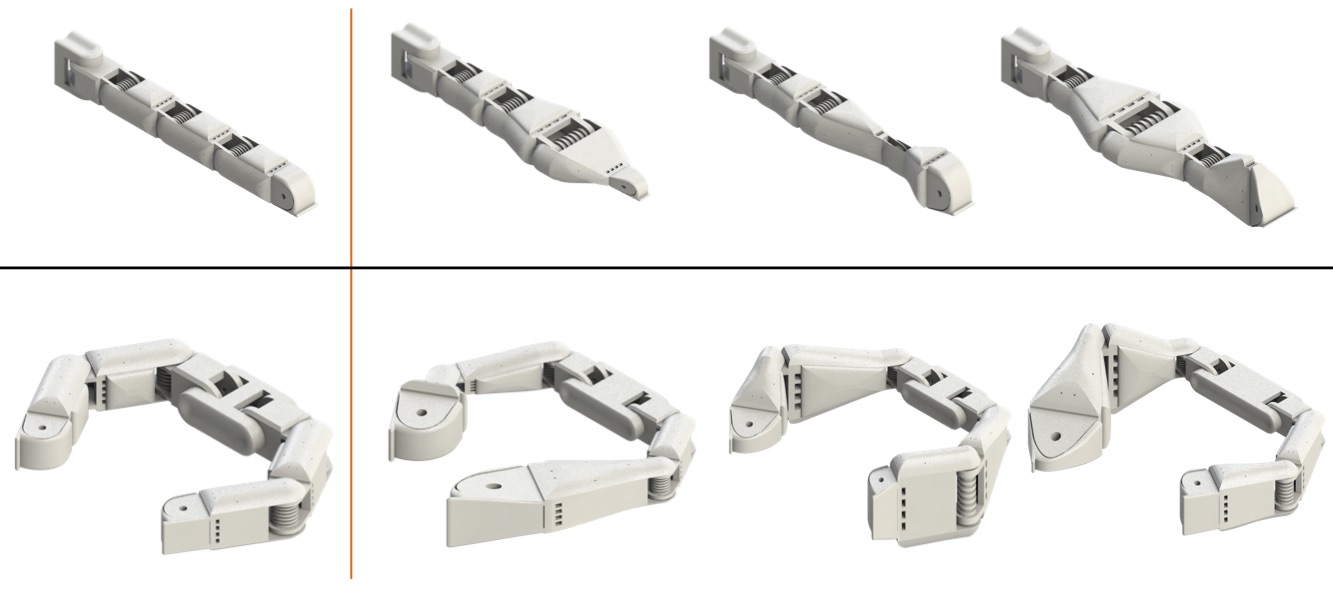}
    
    \caption{\textbf{Morphology design space}. The initial morphology of the single finger and the two-finger gripper designs are shown on the left. We randomly sample different parameters for each configuration and show the deformed morphology on the right.}
    \label{fig:gallery}
    \vspace{-0.5em}
\end{figure}

To demonstrate the quality of the morphology design space represented by our deformation-based parameterization, we randomly sample the high-level morphology parameters $\psi_c$ for two manipulator topologies: \textit{single-finger} and \textit{two-finger gripper} configurations (Figure \ref{fig:gallery}). Our deformation-based parameterization method has a compact, yet expressive and rich design space. As shown in Figure \ref{fig:gallery}, using only $9$ and $17$ design parameters respectively for the two configurations, our method is able to generate various natural and fabricable morphology designs of articulated robots while maintaining the connectivity of the articulated manipulators. 

\begin{table}[h!]
\caption{List of hyper-parameters for each example.}
\centering
\label{tab:example_params}
\vspace{-1em}
\begin{tabular}{lrrrrrr}
\toprule
\textbf{Task} & $\Delta t_s$ &$n_t$ & $n_{ctrl}$ & $|u|$ & $|\psi_c|$ & $|\psi_p|$\\ 
\midrule
\midrule
Finger Reach        &$0.005$&$600$&$20$&$120$&  $9$& $376$\\ \midrule
Flip Box            &$0.005$&$150$& $5$&$180$&  $9$&$1478$\\ \midrule
Rotate Rubik's Cube &$0.005$&$200$& $5$&$240$&  $9$&$1478$\\ \midrule
Assemble            &$0.001$&$500$& $5$&$800$& $17$&$1226$\\ \midrule
Free-form Gripper   &$0.005$&$400$& $1$&  -  &$396$&$9228$\\ 
\bottomrule
\end{tabular}
\vspace{-0.5em}
\end{table}

\subsection{Morphology and Control Co-Optimization}
\begin{figure*}[t!]
    \centering
    \includegraphics[width=\linewidth]{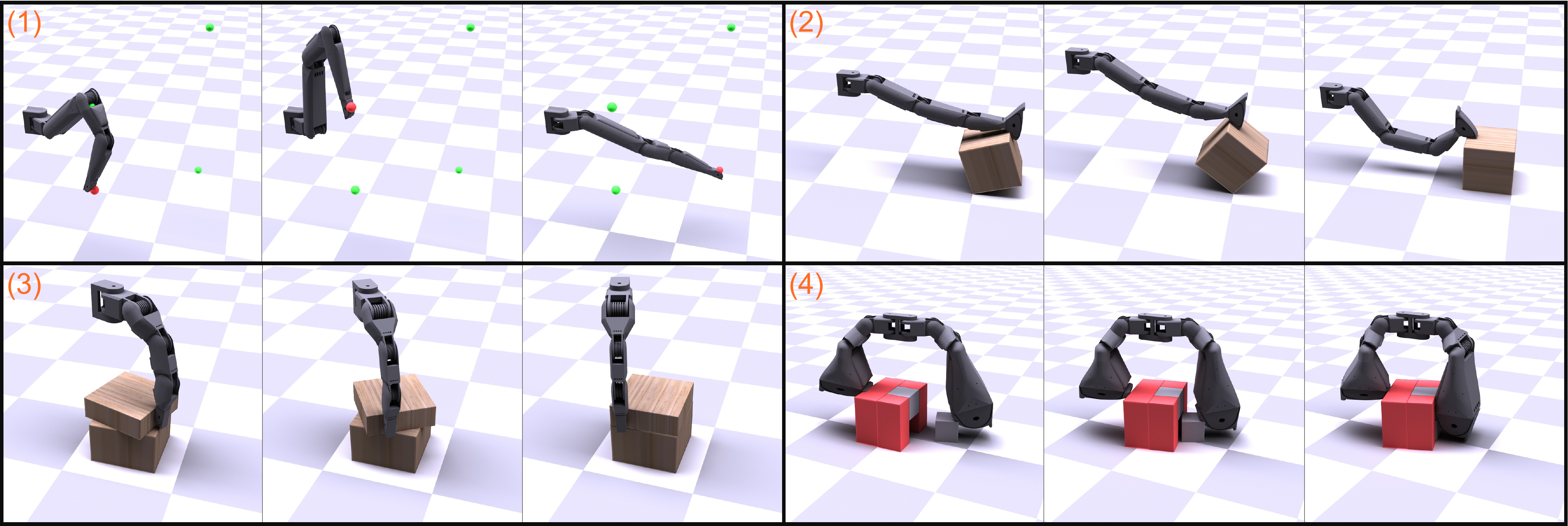}
    \caption{\textbf{Optimized designs and controls for four manipulator tasks}. (1) \textit{Finger Reach}. (2) \textit{Flip Box}. (3) \textit{Rotate Rubik's Cube}. (4) \textit{Assemble}. More visual results are provided in the supplementary video.}
    \label{fig:results_visual}
    \vspace{-4mm}
\end{figure*}

\textbf{Tasks}~
In order to test the performance of our differentiable contact-aware co-optimization framework, we designed four manipulation tasks as shown in Figure \ref{fig:results_visual}, consisting of three single-finger tasks and one two-finger task:
\begin{enumerate}
\item \textit{Finger Reach}: In this task, the base of the finger is mounted on the wall, and the finger is required to reach four scattered target points in the space sequentially. 
\item \textit{Flip Box}: This task requires the finger to flip a heavy box by $90^\circ$ and be as energy-efficient as possible.
\item \textit{Rotate Rubik's Cube}: A finger is required to rotate the top layer of a Rubik's cube by $90^\circ$. The bottom of the Rubik's cube is fixed on the ground.
\item \textit{Assemble}: Two fingers need to collaborate to push and insert a small cube into its movable mount. The cube and the hole on the mount have similar sizes, making the task much more challenging and requiring high-accuracy manipulation.
\end{enumerate}

The detailed description and loss function $\mathcal{L}$ of each task is provided in the Appendix II-A. The hyper-parameters of each task are listed in Table \ref{tab:example_params}, {\change{where $\Delta t_s$ is the simulation time steps size, $n_t$ is the total number of simulation steps of the task, $|u|$ and $|\psi_c|$ are the total numbers of control and morphology variables in optimization respectively, and $|\psi_p|$ is the number of simulation parameters.
We optimize for the control signals not every simulation step but every $n_{ctrl}$ steps, giving us $|u| = (n_t/n_{ctrl})\cdot|u_i|$ where $|u_i|$ is the number of control degrees of freedom of the robot.}}

\begin{figure}[t!]
    \centering
    \includegraphics[width=\linewidth]{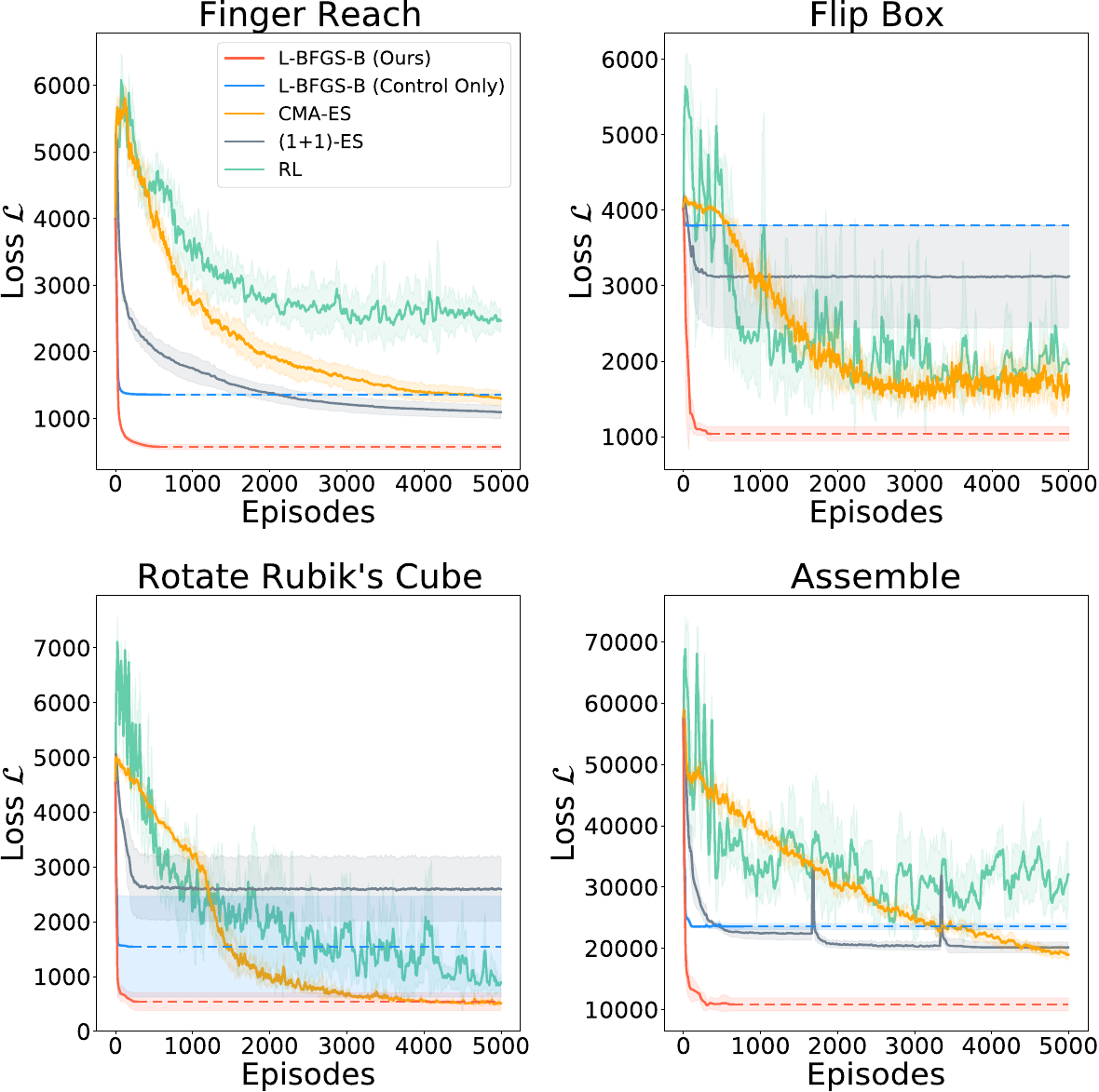}
    \caption{\textbf{Optimization curves comparison}. We run all the methods on all tasks $5$ times with different random seeds. Mean and standard deviation in the loss objective are reported. The horizontal axis of each plot is the number of simulation episodes, and vertical axis is the objective loss value. L-BFGS-B optimization can terminate early once it satisfies the termination criterion. For better visualization, we extend the actual learning curves that use L-BFGS-B horizontally using dotted lines. We also smooth out the curves with a window size of $10$.}
    \label{fig:results_curves}
    \vspace{-0.5em}
\end{figure}

\begin{table*}[h!]
\vspace{-1em}
\caption{Normalized Metric Comparison. \textup{We design the task-related metrics to measure how successful each method performs on the tasks. For \textit{Finger Reach} task, the metric is the time-averaged distance to the target tracking points. For \textit{Flip Box} and \textit{Rotate Rubik's Cube}, the metrics are the flipping/rotating angle error at the end of the task. For \textit{Assemble}, we measure the distance between the center of the small box and the center of the hole on the movable mount. All the metrics are normalized.}}
\centering
\label{tab:quantitative_metric}
\vspace{-1em}
\begin{tabular}{lcccc}
\toprule
\textbf{Task} & Finger Reach & Flip Box & Rotate Rubik's Cube & Assemble \\ 
\midrule
\midrule
CMA-ES &  $0.39 \pm 0.02$  &  $0.00 \pm 0.00$  & $\bm{0.02 \pm 0.01}$ &  $0.28 \pm 0.03$ \\
\midrule
(1+1)-ES &  $0.35 \pm 0.04$  &  $0.69 \pm 0.39$  &  $0.87 \pm 0.15$  &  $0.39 \pm 0.09$ \\
\midrule
RL &  $0.61 \pm 0.05$  &  $0.41 \pm 0.48$  &  $0.79 \pm 0.31$  &  $0.91 \pm 0.11$ \\
\midrule
L-BFGS-B (Control Only) &  $0.41 \pm 0.00$  &  $1.00 \pm 0.00$  &  $0.42 \pm 0.39$  &  $0.77 \pm 0.03$ \\
\midrule
L-BFGS-B (Ours) & $\bm{0.17 \pm 0.01}$ & $\bm{0.00 \pm 0.00}$ &  $0.07 \pm 0.09$  & $\bm{0.12 \pm 0.11}$\\
\bottomrule
\end{tabular}
\end{table*}

\textbf{Baselines}~
We adopted the following three baseline algorithms for comparison.
\begin{enumerate}
\item \textit{ES}: Evolutionary strategy is widely used to search for optimal design and control parameters for robots \cite{digumarti2014concurrent, meixner2019automated}. We tried various ES algorithms in the open-sourced Nevergrad library \cite{nevergrad} and found that the \textit{$(1+1)$-ES}~\cite{auger2009benchmarking, rechenberg1978evolutionsstrategien} algorithm and \textit{CMA-ES}~\cite{hansen2001completely} work best on the proposed tasks. 
\item \textit{RL}: \citet{luck2020data} is one of the state-of-the-art morphology and control co-optimization approaches using sample-efficient reinforcement learning (soft actor critic, SAC) algorithm and particle swarm optimization. We used their released implementation as a baseline.
\item \textit{Control Only}: In this algorithm, we freeze the morphology parameters and only optimize the control sequence with L-BFGS-B.
\end{enumerate}

\textbf{Experiment Setup}~
We use the same morphology parameterization for baselines and our method. The control parameter for the \textit{RL} baseline is a neural network controller (a policy network) as proposed by \citet{luck2020data}, and is an open-loop control sequence for all other methods. We try both \textit{ES} algorithms due to their different performances on different tasks. For fair comparison, we finetune the parameters of \textit{ES} and \textit{RL} baselines and run the experiments with the best-performing parameters. For the \textit{Control Only} baseline and our method, we use the default parameters provided in the Scipy's L-BFGS-B optimizer. While our method can solve the \textit{Assemble} task with a high success rate, we found that the loss objective can be further decreased by using a continuation method \cite{allgower2003continuation,geilinger2020add}. Specifically, on the \textit{Assemble} task, we scale down the contact forces at the beginning of the optimization to provide a smoother objective function space, and scale it up as the optimization proceeds. We set three stages with contact force scale equal to $0.01$, $0.1$ and $1$, and start the next optimization stage once the previous stage converges. To apply the continuation method on the baseline algorithms, we fix the number of simulations in each stage and proceed to the next stage after the previous stage uses up the budget. We run all the baseline algorithms with and without continuation method on \textit{Assemble} task, and plot the best performing one. We run all the methods on each task for five times with different random seeds and plot the average training curves in Figure \ref{fig:results_curves}. We further measure the successfulness of the tasks for each method by task-related metrics, and report in the Table \ref{tab:quantitative_metric}.

\textbf{Results}~
The results show that our differentiable co-optimization framework is able to find better morphology and control solutions with significant better sample efficiency (10-30 times fewer simulated episode data) compared to the gradient-free \textit{ES} baselines and model-free \textit{RL} baseline. On \textit{Finger Reach} task, while most methods (except \textit{Control Only}) find a finger configuration that can reach the four target points, our method can find a morphology and a control sequence that can track the target points most accurately. On the most challenging and contact-rich task, \textit{Assemble}, our method is the only one that is able to solve the task successfully. 

We also performed an ablation study on the importance of morphology design by comparing the performance of our method and a \textit{Control Only} baseline. The significant performance gain of our method over \textit{Control Only} baseline reveals that incorporating the optimization of morphology design leads to easier optimization and better solutions. We show some of the optimized morphology designs from our method in Figure \ref{fig:results_visual}. On the \textit{Flip Box} task, the optimized morphology has a hook-like structure at the finger tip, so that it is able to hook on the back surface of the box to flip over the box more easily. For the \textit{Assemble} task, the optimized morphology has fingers of different lengths, so that the long right finger is able to push the smaller cube while the short left finger can hold the mount. Moreover, the design has flat and larger fingers, which allows the manipulator to push the object much more stably than a thin finger. More visual results can be found in our supplementary video.

\subsection{Flexibility of the Morphology Parameterization}
\begin{figure}[t!]
    \centering
    \includegraphics[width=\linewidth, height=5cm]{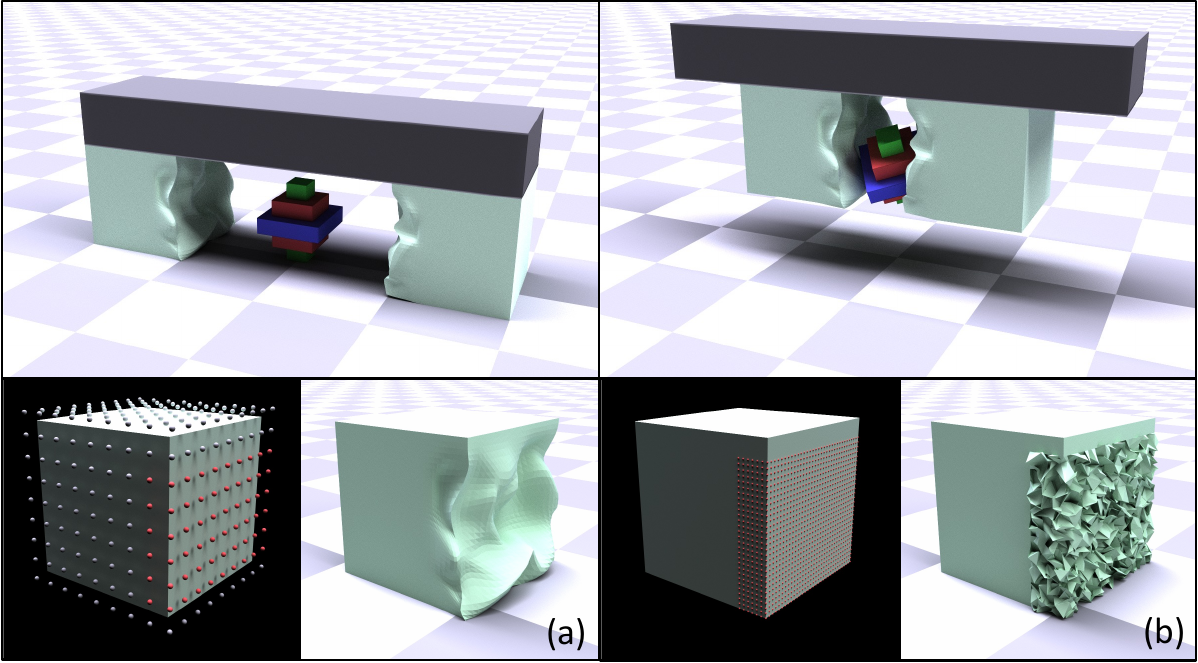}
    \caption{\textbf{Free-form Gripper}: The task is to pick up the object, as shown in the top row. We compare deformation-based parameterization (ours) and mesh-based parameterization. The optimization variables and optimized gripper morphology for the left gripper finger using both methods are shown in the bottom row. (a) \textit{our parameterization method}: all the cage handles are shown on the left sub-figure and the ones used as optimization variables are highlighted in red. (b) \textit{mesh-based parameterization}: we allow the optimization to directly optimize all the mesh vertices highlighted in red in the left sub-figure. In both cases, we do not modify the areas near the top of the gripper. The gripper morphology generated by our method is much smoother.}
    \label{fig:freeform_gripper}
    \vspace{-2mm}
\end{figure}

By adding more cage handles, we can easily increase the degrees of freedom for the morphology design. We show such flexibility of our deformation-based morphology parameterization in this section using a free-form gripper task motivated by \citet{ha2020fit2form}. As shown in Figure \ref{fig:freeform_gripper}, the algorithm needs to optimize the shape of a pair of gripper fingers such that the gripper can grasp a diamond-like object using a predefined control sequence. Each finger starts with a cube-like shape and is optimized with free-form deformation. 

To support deformation in higher degrees of freedom (DoF), we add more handles on the cage around each gripper finger. To show the advantage of using a deformation cage based parameterization, we compare it to a differentiable mesh-based parameterization (similar to the Truncated Signed Distance Function (TSDF) parameterization used in \citet{ha2020fit2form}) which directly optimize over the vertex positions of the mesh. For both parameterization methods, we only optimize the handles/mesh vertices on the inner side for each gripper finger as shown in Figure \ref{fig:freeform_gripper} (bottom row). We test both parameterization methods in our differentiable framework, and conduct $30$ independent experiments for each method with different initial parameters.  

Even though our deformation cage based parameterization has a much smaller optimization space and DoFs ($396$ optimization variables in the grasping task) than the mesh-based parameterization ($8946$ variables), our method is able to generate comparable success rates on the grasping task than the mesh-based parameterization (ours: $97\%$, mesh: $100\%$), and achieves better average loss (plot shown in Appendix II-B). Moreover, as shown in the Figure \ref{fig:freeform_gripper}, our method is able to generate much smoother morphologies than the mesh parameterization which creates many reverted triangles on the mesh. Such advantage comes from the smoothness and feature-preserving properties of the cage-based deformation method.

\subsection{\change{Manufacturing of Optimized Designs}}
We manufactured two optimized finger models from our method, one generated by the \textit{Flip Box} task and the other generated by the \textit{Assemble} task, as shown in Figure \ref{fig:intro}. The finger components were 3D printed on a Markforged printer using Onyx, a micro carbon fiber filled nylon, and assembled together after print. Minimal modification was required to prepare the program-generated models for printing, demonstrating a streamlined design process. 
This shows that our deformation-based morphology paramterization successfully maintains design manufacturability.

{\change{We further tested the functionality of the manufactured finger for the \textit{Flip Box} task. Vectran cables were routed through the 3D printed finger and controlled by dynamixel DC servos at the base of the finger. The tendon-driven finger system was then mounted on a UR5 arm. We manually programmed a control sequence on the dynamixel motors and the UR5 arm to follow a series of waypoints from the trajectory optimized by the algorithm. The experiment shows that the manufactured finger can effectively flip the cube in real world. We also test the robustness of the optimized design on the cubes of various sizes and the experiment demonstrates that the finger can also perform the flipping tasks successfully. Please see the supplementary video for this real-world experiment.}}

\section{Discussion}
In this work, we present an end-to-end differentiable framework for contact-aware robot designs. At the core of our contribution is a novel deformation-based morphology parameterization for articulated robot designs, and a differentiable rigid body simulation carefully developed for contact-rich manipulation tasks. The experiments show that our innovative morphology parameterization approach provides us with an effective and expressive morphology design space. We also demonstrate that for a given manipulation task, by applying gradient-based optimization algorithm in our fully differentiable framework, our method is able to find a better morphology and control combination with significantly fewer number of simulation episodes than the state-of-the-art approaches. {\change{Furthermore, the optimized designs can be easily manufactured and are functional in real world.}} We release our implementation for reproducibility. 

{\change{Note that although in this work we focus on optimizing the continuous component shape given a fixed design topology, the database we constructed can actually produce a large amount of different design topologies (\eg different number of fingers, different number of links, etc.) via different combinations of the components through our constructed grammar in Appendix I. Furthermore, our deformation-based parameterization allows us to reuse the cages and the precomputed deformation weights for each individual component across different manipulator structures. Such automatic and repeatable process enables the potential research on manipulator structure optimization in the future.}}
{\change{It is also worth mentioning that while our examples only show the components with identical connection surfaces, it is not a limitation of our proposed deformation-based parameterization. One can easily applies our method for components with connection surfaces of different sizes by constructing proper cages for them.}}

There are several directions which can be explored in the future. First, though analytical derivatives allows for an efficient gradient-based optimization algorithm, the local minimum issue of the gradient-based optimization is a known problem. Combining the exploration ability from model-free (gradient-free) methods and the gradient-based optimization can be an interesting direction to better leverage the advantages from both worlds. Second, our proposed method is applicable to optimize the continuous morphology parameters given the fixed robot topology. There is another category of morphology optimization that focuses on searching the discrete topology of the robot designs \cite{ha2018computational, wang2019neural, xu2021moghs, zhao2020robogrammar}. It is desirable to develop an advanced optimization algorithm to effectively search over both the discrete robot topology space and the continuous morphology space. Finally, narrowing down the gap between the reality and our simulation-based framework can be another important task in the future in order to seamlessly transfer the optimized morphology and control from simulation directly in to real world.

\section*{Acknowledgements}
We thank the anonymous reviewers for their helpful comments in revising the paper. Toyota Research Institute (TRI), Defense Advanced Research Projects Agency (FA8750-20-C-0075), and the National Science Foundation (CAREER-1846368) provided funds to support this work.


\bibliographystyle{plainnat}
\bibliography{main}

\begin{thebibliography}{55}
\providecommand{\natexlab}[1]{#1}
\providecommand{\url}[1]{\texttt{#1}}
\expandafter\ifx\csname urlstyle\endcsname\relax
  \providecommand{\doi}[1]{doi: #1}\else
  \providecommand{\doi}{doi: \begingroup \urlstyle{rm}\Url}\fi

\bibitem[Akkaya et~al.(2019)Akkaya, Andrychowicz, Chociej, Litwin, McGrew,
  Petron, Paino, Plappert, Powell, Ribas, et~al.]{akkaya2019solving}
Ilge Akkaya, Marcin Andrychowicz, Maciek Chociej, Mateusz Litwin, Bob McGrew,
  Arthur Petron, Alex Paino, Matthias Plappert, Glenn Powell, Raphael Ribas,
  et~al.
\newblock Solving rubik's cube with a robot hand.
\newblock \emph{arXiv preprint arXiv:1910.07113}, 2019.

\bibitem[Allgower and Georg(2003)]{allgower2003continuation}
Eugene~L Allgower and Kurt Georg.
\newblock \emph{Introduction to numerical continuation methods}.
\newblock SIAM, 2003.

\bibitem[Amos and Kolter(2019)]{amos2019optnet}
Brandon Amos and J.~Zico Kolter.
\newblock Optnet: Differentiable optimization as a layer in neural networks,
  2019.

\bibitem[Andrychowicz et~al.(2020)Andrychowicz, Baker, Chociej, Jozefowicz,
  McGrew, Pachocki, Petron, Plappert, Powell, Ray,
  et~al.]{andrychowicz2020learning}
OpenAI:~Marcin Andrychowicz, Bowen Baker, Maciek Chociej, Rafal Jozefowicz, Bob
  McGrew, Jakub Pachocki, Arthur Petron, Matthias Plappert, Glenn Powell, Alex
  Ray, et~al.
\newblock Learning dexterous in-hand manipulation.
\newblock \emph{The International Journal of Robotics Research}, 39\penalty0
  (1):\penalty0 3--20, 2020.

\bibitem[Auger(2009)]{auger2009benchmarking}
Anne Auger.
\newblock Benchmarking the (1+ 1) evolution strategy with one-fifth success
  rule on the bbob-2009 function testbed.
\newblock In \emph{Proceedings of the 11th Annual Conference Companion on
  Genetic and Evolutionary Computation Conference: Late Breaking Papers}, pages
  2447--2452, 2009.

\bibitem[Battaglia et~al.(2016)Battaglia, Pascanu, Lai, Rezende, and
  kavukcuoglu]{battaglia2016interaction}
Peter Battaglia, Razvan Pascanu, Matthew Lai, Danilo~Jimenez Rezende, and Koray
  kavukcuoglu.
\newblock Interaction networks for learning about objects, relations and
  physics.
\newblock In \emph{Proceedings of the 30th International Conference on Neural
  Information Processing Systems}, pages 4509--4517, 2016.

\bibitem[Brockman et~al.(2016)Brockman, Cheung, Pettersson, Schneider,
  Schulman, Tang, and Zaremba]{brockman2016openai}
Greg Brockman, Vicki Cheung, Ludwig Pettersson, Jonas Schneider, John Schulman,
  Jie Tang, and Wojciech Zaremba.
\newblock Openai gym.
\newblock \emph{arXiv preprint arXiv:1606.01540}, 2016.

\bibitem[Chang et~al.(2016)Chang, Ullman, Torralba, and
  Tenenbaum]{chang2016compositional}
Michael~B Chang, Tomer Ullman, Antonio Torralba, and Joshua~B Tenenbaum.
\newblock A compositional object-based approach to learning physical dynamics.
\newblock \emph{arXiv preprint arXiv:1612.00341}, 2016.

\bibitem[Chen et~al.(2020)Chen, He, and Ciocarlie]{chen2020hardware}
Tianjian Chen, Zhanpeng He, and Matei Ciocarlie.
\newblock Hardware as policy: Mechanical and computational co-optimization
  using deep reinforcement learning.
\newblock \emph{arXiv preprint arXiv:2008.04460}, 2020.

\bibitem[Coumans and Bai(2016)]{coumans2016pybullet}
Erwin Coumans and Yunfei Bai.
\newblock Pybullet, a python module for physics simulation for games, robotics
  and machine learning.
\newblock 2016.

\bibitem[de~Avila Belbute-Peres et~al.(2018)de~Avila Belbute-Peres, Smith,
  Allen, Tenenbaum, and Kolter]{de2018end}
Filipe de~Avila Belbute-Peres, Kevin Smith, Kelsey Allen, Josh Tenenbaum, and
  J~Zico Kolter.
\newblock End-to-end differentiable physics for learning and control.
\newblock \emph{Advances in neural information processing systems},
  31:\penalty0 7178--7189, 2018.

\bibitem[Degrave et~al.(2019)Degrave, Hermans, Dambre,
  et~al.]{degrave2019differentiable}
Jonas Degrave, Michiel Hermans, Joni Dambre, et~al.
\newblock A differentiable physics engine for deep learning in robotics.
\newblock \emph{Frontiers in neurorobotics}, 13:\penalty0 6, 2019.

\bibitem[Deimel and Brock(2016)]{deimel2016novel}
Raphael Deimel and Oliver Brock.
\newblock A novel type of compliant and underactuated robotic hand for
  dexterous grasping.
\newblock \emph{The International Journal of Robotics Research}, 35\penalty0
  (1-3):\penalty0 161--185, 2016.

\bibitem[Deimel et~al.(2017)Deimel, Irmisch, Wall, and Brock]{8202294}
Raphael Deimel, Patrick Irmisch, Vincent Wall, and Oliver Brock.
\newblock Automated co-design of soft hand morphology and control strategy for
  grasping.
\newblock In \emph{2017 IEEE/RSJ International Conference on Intelligent Robots
  and Systems (IROS)}, pages 1213--1218, 2017.
\newblock \doi{10.1109/IROS.2017.8202294}.

\bibitem[Digumarti et~al.(2014)Digumarti, Gehring, Coros, Hwangbo, and
  Siegwart]{digumarti2014concurrent}
KM~Digumarti, C~Gehring, S~Coros, J~Hwangbo, and R~Siegwart.
\newblock Concurrent optimization of mechanical design and locomotion control
  of a legged robot.
\newblock In \emph{17th International Conference on Climbing and Walking Robots
  (CLAWAR)}, pages 315--+. WORLD SCIENTIFIC PUBL CO PTE LTD, 2014.

\bibitem[Du et~al.(2021)Du, Wu, Ma, Wah, Spielberg, Rus, and
  Matusik]{du2021diffpd}
Tao Du, Kui Wu, Pingchuan Ma, Sebastien Wah, Andrew Spielberg, Daniela Rus, and
  Wojciech Matusik.
\newblock Diffpd: Differentiable projective dynamics with contact.
\newblock \emph{arXiv preprint arXiv:2101.05917}, 2021.

\bibitem[Geilinger et~al.(2020)Geilinger, Hahn, Zehnder, B{\"a}cher,
  Thomaszewski, and Coros]{geilinger2020add}
Moritz Geilinger, David Hahn, Jonas Zehnder, Moritz B{\"a}cher, Bernhard
  Thomaszewski, and Stelian Coros.
\newblock Add: analytically differentiable dynamics for multi-body systems with
  frictional contact.
\newblock \emph{ACM Transactions on Graphics (TOG)}, 39\penalty0 (6):\penalty0
  1--15, 2020.

\bibitem[Gupta et~al.(2021)Gupta, Savarese, Ganguli, and
  Fei-Fei]{gupta2021embodied}
Agrim Gupta, Silvio Savarese, Surya Ganguli, and Li~Fei-Fei.
\newblock Embodied intelligence via learning and evolution.
\newblock \emph{arXiv preprint arXiv:2102.02202}, 2021.

\bibitem[Ha et~al.(2020)Ha, Agrawal, and Song]{ha2020fit2form}
Huy Ha, Shubham Agrawal, and Shuran Song.
\newblock Fit2form: 3d generative model for robot gripper form design.
\newblock \emph{arXiv preprint arXiv:2011.06498}, 2020.

\bibitem[Ha et~al.(2017)Ha, Coros, Alspach, Kim, and Yamane]{ha2017joint}
Sehoon Ha, Stelian Coros, Alexander Alspach, Joohyung Kim, and Katsu Yamane.
\newblock Joint optimization of robot design and motion parameters using the
  implicit function theorem.
\newblock In \emph{Robotics: Science and systems}, volume~8, 2017.

\bibitem[Ha et~al.(2018)Ha, Coros, Alspach, Bern, Kim, and
  Yamane]{ha2018computational}
Sehoon Ha, Stelian Coros, Alexander Alspach, James~M Bern, Joohyung Kim, and
  Katsu Yamane.
\newblock Computational design of robotic devices from high-level motion
  specifications.
\newblock \emph{IEEE Transactions on Robotics}, 34\penalty0 (5):\penalty0
  1240--1251, 2018.

\bibitem[Hafner et~al.(2019)Hafner, Schumacher, Knoop, Auzinger, Bickel, and
  B\"{a}cher]{10.1145/3355089.3356576}
Christian Hafner, Christian Schumacher, Espen Knoop, Thomas Auzinger, Bernd
  Bickel, and Moritz B\"{a}cher.
\newblock X-cad: Optimizing cad models with extended finite elements.
\newblock \emph{ACM Trans. Graph.}, 38\penalty0 (6), November 2019.
\newblock ISSN 0730-0301.
\newblock \doi{10.1145/3355089.3356576}.
\newblock URL \url{https://doi.org/10.1145/3355089.3356576}.

\bibitem[Hahn et~al.(2019)Hahn, Banzet, Bern, and Coros]{hahn2019real2sim}
David Hahn, Pol Banzet, James~M Bern, and Stelian Coros.
\newblock Real2sim: Visco-elastic parameter estimation from dynamic motion.
\newblock \emph{ACM Transactions on Graphics (TOG)}, 38\penalty0 (6):\penalty0
  1--13, 2019.

\bibitem[Hansen and Ostermeier(2001)]{hansen2001completely}
Nikolaus Hansen and Andreas Ostermeier.
\newblock Completely derandomized self-adaptation in evolution strategies.
\newblock \emph{Evolutionary computation}, 9\penalty0 (2):\penalty0 159--195,
  2001.

\bibitem[Hazard et~al.(2018)Hazard, Pollard, and Coros]{hazard2018automated}
Christopher Hazard, Nancy Pollard, and Stelian Coros.
\newblock Automated design of manipulators for in-hand tasks.
\newblock In \emph{2018 IEEE-RAS 18th International Conference on Humanoid
  Robots (Humanoids)}, pages 1--8. IEEE, 2018.

\bibitem[Heiden et~al.(2019)Heiden, Millard, Zhang, and
  Sukhatme]{heiden2019interactive}
Eric Heiden, David Millard, Hejia Zhang, and Gaurav~S Sukhatme.
\newblock Interactive differentiable simulation.
\newblock \emph{arXiv preprint arXiv:1905.10706}, 2019.

\bibitem[Hu et~al.(2019)Hu, Liu, Spielberg, Tenenbaum, Freeman, Wu, Rus, and
  Matusik]{hu2019chainqueen}
Yuanming Hu, Jiancheng Liu, Andrew Spielberg, Joshua~B Tenenbaum, William~T
  Freeman, Jiajun Wu, Daniela Rus, and Wojciech Matusik.
\newblock Chainqueen: A real-time differentiable physical simulator for soft
  robotics.
\newblock In \emph{2019 International conference on robotics and automation
  (ICRA)}, pages 6265--6271. IEEE, 2019.

\bibitem[Hu et~al.(2020)Hu, Anderson, Li, Sun, Carr, Ragan-Kelley, and
  Durand]{hu2020difftaichi}
Yuanming Hu, Luke Anderson, Tzu-Mao Li, Qi~Sun, Nathan Carr, Jonathan
  Ragan-Kelley, and Fr{\'e}do Durand.
\newblock Difftaichi: Differentiable programming for physical simulation.
\newblock In \emph{International Conference on Learning Representations}, 2020.

\bibitem[Huang et~al.(2021)Huang, Hu, Du, Zhou, Su, Tenenbaum, and
  Gan]{huang2021plasticinelab}
Zhiao Huang, Yuanming Hu, Tao Du, Siyuan Zhou, Hao Su, Joshua~B. Tenenbaum, and
  Chuang Gan.
\newblock Plasticinelab: A soft-body manipulation benchmark with differentiable
  physics.
\newblock In \emph{International Conference on Learning Representations}, 2021.
\newblock URL \url{https://openreview.net/forum?id=xCcdBRQEDW}.

\bibitem[Jacobson et~al.(2011)Jacobson, Baran, Popovic, and
  Sorkine]{jacobson2011bounded}
Alec Jacobson, Ilya Baran, Jovan Popovic, and Olga Sorkine.
\newblock Bounded biharmonic weights for real-time deformation.
\newblock \emph{ACM Trans. Graph.}, 30\penalty0 (4):\penalty0 78, 2011.

\bibitem[Jatavallabhula et~al.(2021)Jatavallabhula, Macklin, Golemo, Voleti,
  Petrini, Weiss, Considine, Parent-Levesque, Xie, Erleben,
  et~al.]{jatavallabhula2021gradsim}
Krishna~Murthy Jatavallabhula, Miles Macklin, Florian Golemo, Vikram Voleti,
  Linda Petrini, Martin Weiss, Breandan Considine, Jerome Parent-Levesque,
  Kevin Xie, Kenny Erleben, et~al.
\newblock gradsim: Differentiable simulation for system identification and
  visuomotor control.
\newblock \emph{arXiv preprint arXiv:2104.02646}, 2021.

\bibitem[Ju et~al.(2005)Ju, Schaefer, and Warren]{10.1145/1186822.1073229}
Tao Ju, Scott Schaefer, and Joe Warren.
\newblock Mean value coordinates for closed triangular meshes.
\newblock In \emph{ACM SIGGRAPH 2005 Papers}, SIGGRAPH '05, page 561–566, New
  York, NY, USA, 2005. Association for Computing Machinery.
\newblock ISBN 9781450378253.
\newblock \doi{10.1145/1186822.1073229}.
\newblock URL \url{https://doi.org/10.1145/1186822.1073229}.

\bibitem[Li et~al.(2018)Li, Wu, Tedrake, Tenenbaum, and
  Torralba]{li2018learning}
Yunzhu Li, Jiajun Wu, Russ Tedrake, Joshua~B Tenenbaum, and Antonio Torralba.
\newblock Learning particle dynamics for manipulating rigid bodies, deformable
  objects, and fluids.
\newblock \emph{arXiv preprint arXiv:1810.01566}, 2018.

\bibitem[Liang et~al.(2019)Liang, Lin, and Koltun]{NEURIPS2019_28f0b864}
Junbang Liang, Ming Lin, and Vladlen Koltun.
\newblock Differentiable cloth simulation for inverse problems.
\newblock In H.~Wallach, H.~Larochelle, A.~Beygelzimer, F.~d\textquotesingle
  Alch\'{e}-Buc, E.~Fox, and R.~Garnett, editors, \emph{Advances in Neural
  Information Processing Systems}, volume~32. Curran Associates, Inc., 2019.
\newblock URL
  \url{https://proceedings.neurips.cc/paper/2019/file/28f0b864598a1291557bed248a998d4e-Paper.pdf}.

\bibitem[Luck et~al.(2020)Luck, Amor, and Calandra]{luck2020data}
Kevin~Sebastian Luck, Heni~Ben Amor, and Roberto Calandra.
\newblock Data-efficient co-adaptation of morphology and behaviour with deep
  reinforcement learning.
\newblock In \emph{Conference on Robot Learning}, pages 854--869. PMLR, 2020.

\bibitem[McNamara et~al.(2004)McNamara, Treuille, Popovi\'{c}, and
  Stam]{McNamara2004}
Antoine McNamara, Adrien Treuille, Zoran Popovi\'{c}, and Jos Stam.
\newblock Fluid control using the adjoint method.
\newblock \emph{ACM Trans. Graph.}, 23\penalty0 (3):\penalty0 449–456, August
  2004.
\newblock ISSN 0730-0301.
\newblock \doi{10.1145/1015706.1015744}.
\newblock URL \url{https://doi.org/10.1145/1015706.1015744}.

\bibitem[Meixner et~al.(2019)Meixner, Hazard, and
  Pollard]{meixner2019automated}
Andre Meixner, Christopher Hazard, and Nancy Pollard.
\newblock Automated design of simple and robust manipulators for dexterous
  in-hand manipulation tasks using evolutionary strategies.
\newblock In \emph{2019 IEEE-RAS 19th International Conference on Humanoid
  Robots (Humanoids)}, pages 281--288. IEEE, 2019.

\bibitem[Mrowca et~al.(2018)Mrowca, Zhuang, Wang, Haber, Fei-Fei, Tenenbaum,
  and Yamins]{mrowca2018flexible}
Damian Mrowca, Chengxu Zhuang, Elias Wang, Nick Haber, Li~Fei-Fei, Joshua~B
  Tenenbaum, and Daniel~LK Yamins.
\newblock Flexible neural representation for physics prediction.
\newblock \emph{arXiv preprint arXiv:1806.08047}, 2018.

\bibitem[Nagabandi et~al.(2020)Nagabandi, Konolige, Levine, and
  Kumar]{nagabandi2020deep}
Anusha Nagabandi, Kurt Konolige, Sergey Levine, and Vikash Kumar.
\newblock Deep dynamics models for learning dexterous manipulation.
\newblock In \emph{Conference on Robot Learning}, pages 1101--1112. PMLR, 2020.

\bibitem[Nishikawa(2019)]{Nishikawa2019}
Hiroaki Nishikawa.
\newblock On large start-up error of bdf2.
\newblock 392:\penalty0 456 -- 461, 2019.
\newblock ISSN 0021-9991.

\bibitem[Nocedal and Wright(2006)]{nocedal2006numerical}
Jorge Nocedal and Stephen Wright.
\newblock \emph{Numerical optimization}.
\newblock Springer Science \& Business Media, 2006.

\bibitem[Pan et~al.(2020)Pan, Garg, Anandkumar, and Zhu]{pan2020emergent}
Xinlei Pan, Animesh Garg, Animashree Anandkumar, and Yuke Zhu.
\newblock Emergent hand morphology and control from optimizing robust grasps of
  diverse objects.
\newblock \emph{arXiv preprint arXiv:2012.12209}, 2020.

\bibitem[Qiao et~al.(2020)Qiao, Liang, Koltun, and Lin]{qiao2020scalable}
Yi-Ling Qiao, Junbang Liang, Vladlen Koltun, and Ming Lin.
\newblock Scalable differentiable physics for learning and control.
\newblock In \emph{International Conference on Machine Learning}, pages
  7847--7856. PMLR, 2020.

\bibitem[Rapin and Teytaud(2018)]{nevergrad}
J.~Rapin and O.~Teytaud.
\newblock {Nevergrad - A gradient-free optimization platform}.
\newblock \url{https://GitHub.com/FacebookResearch/Nevergrad}, 2018.

\bibitem[Rechenberg(1978)]{rechenberg1978evolutionsstrategien}
Ingo Rechenberg.
\newblock Evolutionsstrategien.
\newblock In \emph{Simulationsmethoden in der Medizin und Biologie}, pages
  83--114. Springer, 1978.

\bibitem[Rus and Tolley(2015)]{rus2015design}
Daniela Rus and Michael~T Tolley.
\newblock Design, fabrication and control of soft robots.
\newblock \emph{Nature}, 521\penalty0 (7553):\penalty0 467--475, 2015.

\bibitem[Schaff et~al.(2019)Schaff, Yunis, Chakrabarti, and
  Walter]{schaff2019jointly}
Charles Schaff, David Yunis, Ayan Chakrabarti, and Matthew~R Walter.
\newblock Jointly learning to construct and control agents using deep
  reinforcement learning.
\newblock In \emph{2019 International Conference on Robotics and Automation
  (ICRA)}, pages 9798--9805. IEEE, 2019.

\bibitem[Schulz et~al.(2017)Schulz, Xu, Zhu, Zheng, Grinspun, and
  Matusik]{schulz2017interactive}
Adriana Schulz, Jie Xu, Bo~Zhu, Changxi Zheng, Eitan Grinspun, and Wojciech
  Matusik.
\newblock Interactive design space exploration and optimization for cad models.
\newblock \emph{ACM Transactions on Graphics (TOG)}, 36\penalty0 (4):\penalty0
  1--14, 2017.

\bibitem[Spielberg et~al.(2017)Spielberg, Araki, Sung, Tedrake, and
  Rus]{spielberg2017functional}
Andrew Spielberg, Brandon Araki, Cynthia Sung, Russ Tedrake, and Daniela Rus.
\newblock Functional co-optimization of articulated robots.
\newblock In \emph{2017 IEEE International Conference on Robotics and
  Automation (ICRA)}, pages 5035--5042. IEEE, 2017.

\bibitem[Todorov et~al.(2012)Todorov, Erez, and Tassa]{todorov2012mujoco}
Emanuel Todorov, Tom Erez, and Yuval Tassa.
\newblock Mujoco: A physics engine for model-based control.
\newblock In \emph{2012 IEEE/RSJ International Conference on Intelligent Robots
  and Systems}, pages 5026--5033. IEEE, 2012.

\bibitem[Wampler and Popovi{\'c}(2009)]{wampler2009optimal}
Kevin Wampler and Zoran Popovi{\'c}.
\newblock Optimal gait and form for animal locomotion.
\newblock \emph{ACM Transactions on Graphics (TOG)}, 28\penalty0 (3):\penalty0
  1--8, 2009.

\bibitem[Wang et~al.(2019{\natexlab{a}})Wang, Zhou, Fidler, and
  Ba]{wang2019neural}
Tingwu Wang, Yuhao Zhou, Sanja Fidler, and Jimmy Ba.
\newblock Neural graph evolution: Towards efficient automatic robot design.
\newblock \emph{arXiv preprint arXiv:1906.05370}, 2019{\natexlab{a}}.

\bibitem[Wang et~al.(2019{\natexlab{b}})Wang, Weidner, Baxter, Hwang, Kaufman,
  and Sueda]{Wang2019}
Ying Wang, Nicholas~J. Weidner, Margaret~A. Baxter, Yura Hwang, Danny~M.
  Kaufman, and Shinjiro Sueda.
\newblock \textsc{RedMax}: Efficient \& flexible approach for articulated
  dynamics.
\newblock \emph{{ACM} Trans.\ Graph.}, 38\penalty0 (4), July
  2019{\natexlab{b}}.
\newblock ISSN 0730-0301.
\newblock \doi{10.1145/3306346.3322952}.
\newblock URL \url{https://doi.org/10.1145/3306346.3322952}.

\bibitem[Xu et~al.(2021)Xu, Speilberg, Zhao, Rus, and Matusik]{xu2021moghs}
Jie Xu, Andrew Speilberg, Allan Zhao, Daniela Rus, and Wojciech Matusik.
\newblock Multi-objective graph heuristic search for terrestrial robot design.
\newblock IEEE, 2021.

\bibitem[Zhao et~al.(2020)Zhao, Xu, Konakovi{\'c}-Lukovi{\'c}, Hughes,
  Spielberg, Rus, and Matusik]{zhao2020robogrammar}
Allan Zhao, Jie Xu, Mina Konakovi{\'c}-Lukovi{\'c}, Josephine Hughes, Andrew
  Spielberg, Daniela Rus, and Wojciech Matusik.
\newblock Robogrammar: graph grammar for terrain-optimized robot design.
\newblock \emph{ACM Transactions on Graphics (TOG)}, 39\penalty0 (6):\penalty0
  1--16, 2020.

\end{thebibliography}

\newpage
\onecolumn
\begin{center}
    \textbf{\Large{Appendix}}
\end{center}

\setcounter{section}{0}
\section{Manipulator Grammar}
While our work focuses on optimizing the continuous morphology parameters for a given robot design topology, we make the full pipeline as automatic as possible, from constructing the design all the way towards the optimization. Inspired by \citet{zhao2020robogrammar}, we represent the manipulator topology as a graph and construct a context-sensitive stochastic graph grammar for the manipulators to generate designs of different topologies and compute their articulated cage parameterization. 

\begin{figure}[h!]
    \centering
    \includegraphics[width=\linewidth]{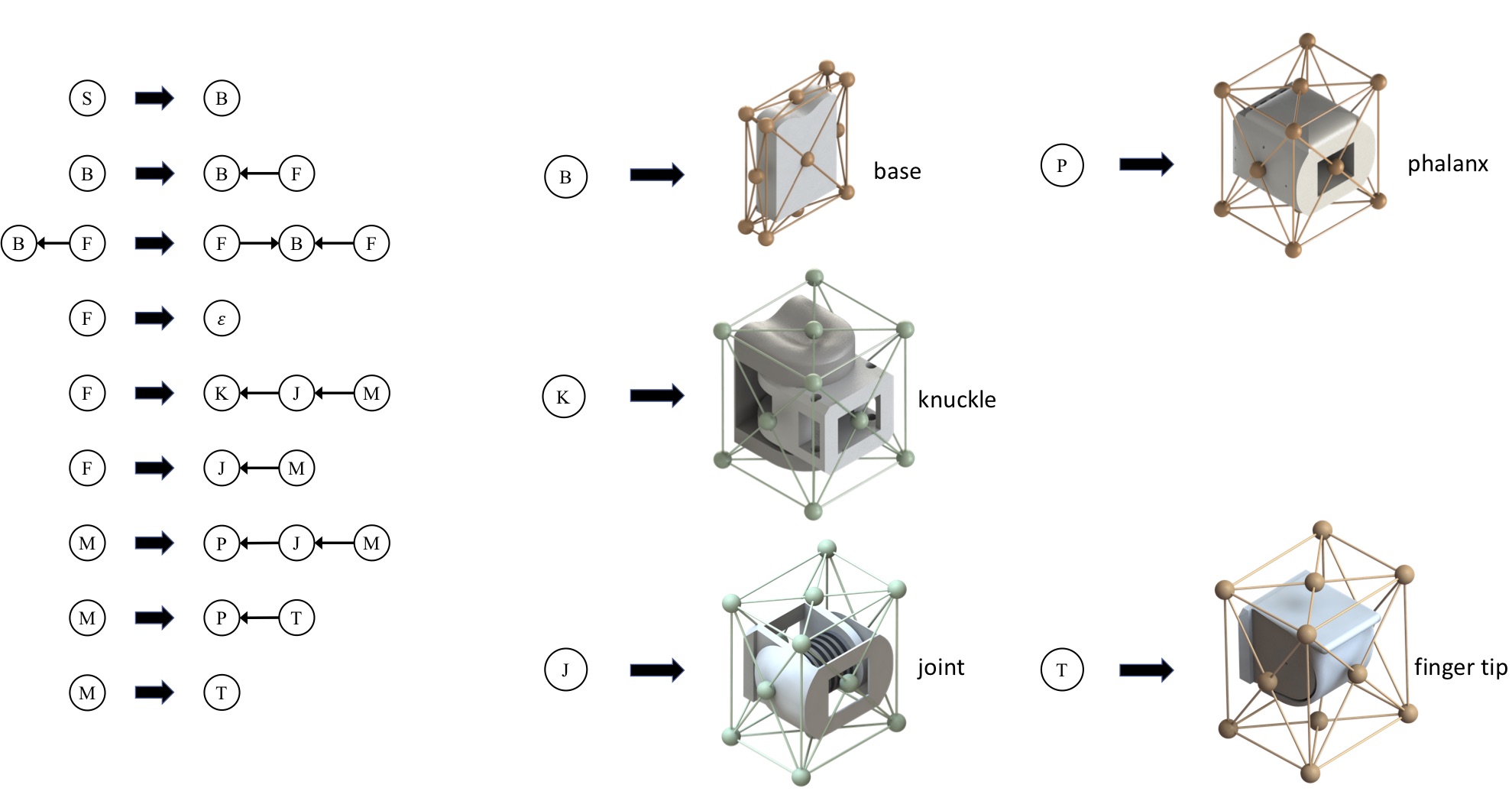}
    
    \caption{\textbf{Context-sensitive stochastic graph grammar for manipulator.}  }
    \label{fig:grammar}
\end{figure}

The grammar is shown in Figure \ref{fig:grammar}. Each rule in the grammar can replace a non-terminal graph node with a sub-graph. More specifically, each rule contains a graph on its left hand side and a graph on its right hand side indicating the sub-graph before and after this rule derivation.  Each circle with a letter is a nonterminal symbol and each one with model picture is a terminal symbol which is also the component from our constructed database. Specially, the circle with $\epsilon$ inside indicates it is an empty node which means that by this rule we are able to remove a nonterminal symbol ``F'' from the graph. To construct a manipulator, we start from the initial nonterminal symbol ``S'' and iteratively apply the rules to replace the nonterminal symbols in the graph until there is no nonterminal symbols in the graph. The grammar system also automatically compute the articulated cages during rule application.

With a small component database consisting of only five basic components, our grammar provides us with a expressive topology design space of various component combinations. The two manipulator configurations are picked among the designs in this grammar space. 

\section{Experiment Details}
In this section, we provide more details about experiments.

\subsection{Tasks}
We elaborate the designed tasks in this section including the task descriptions and the loss functions. We constructed five tasks in the experiments to test the performance of our method.
\begin{enumerate}
    \item \textit{Finger Reach}: In this task, the finger's base is mounted on the wall, and the finger is required to reach four scattered target positions in the space in sequence. The initial design of the finger is not long enough to reach two of them. Thus it requires the algorithm to optimize to finger to be longer in order to reach all four points. The cost $\mathcal{L}$ of this task is computed by:
    \begin{align}
        \mathcal{L} = \sum_{t = 1}^T c_u \|u_t\|^2+c_p\|p_t - \hat{p}_t\| \\
        \text{with } c_u = 0.1, c_p = 10 \nonumber
    \end{align} 
    where $u_t \in [-1, 1]$ is the action at time $t$, $p_t$ is the finger tip position at time $t$, and $\hat{p}_t$ is the target points at time $t$.

    \item \textit{Flip Box}: This task requires the finger to flip a heavy box by $90^\circ$ and be as energy-efficient as possible. The bottom front edge of the heavy box is attached to the ground with a revolute joint. This task is more difficult than the previous one since the finger needs to interact with the box, which involves a rich amount of contacts and requires leverage of the contact force to flip the heavy box. The cost of this task is computed by:
    \begin{align}
        \mathcal{L} = \sum_{t = 1}^T c_u \|u_t\|^2 + c_{touch} \|p_t - p_{touch}\|^2 + c_{flip} \|\theta_t - \frac{\pi}{2}\|^2 \\
        \text{with } c_u = 5, c_{touch} = \left\{\begin{array}{cc}
            1 & t < T / 2 \\
            0 & t \geq T / 2
        \end{array}
        \right., c_{flip} = 50
    \end{align}
    where $u_t \in [-1, 1]$ is the action at time $t$, $p_t$ is the finger tip position at time $t$, $p_{tourch}$ is a point on the back surface of the box, and $\theta_t$ is the rotation angle of the box at time $t$. The second term is designed to encourage the manipulator to touch the box and provide some simple heuristics of solving the task.

    \item \textit{Rotate Rubik's Cube}: The finger is required to rotate the top layer of a Rubik's cube by $90^\circ$. The bottom of the Rubik's cube is fixed on the ground. In this task, there is no clear heuristics in the objective function to guide the finger to touch a specific place on the cube, so the finger needs to be optimized to find the correct strategy. The cost of this task is defined by:
    \begin{align}
        \mathcal{L} = \sum_{t=1}^T\big(c_u\|u_t\|^2 + c_{touch}\|p_t - p_{cube}\|^2\big) + c_{rotate} \|\theta_T - \frac{\pi}{2}\|^2 \\
        \text{with } c_u = 5, c_{touch} = 0.1, c_{rotate} = 1000
    \end{align}
    where $u_t$ and $p_t$ are same as previous tasks, $p_{cube}$ is the center of the Rubik's cube, and $\theta_T$ is the rotation angle of the top layer of the cube at the last time step.

    \item \textit{Assemble}: In this task, two fingers need to collaborate together to push and insert a small box into its movable mount. The box and the hole on the mount have similar sizes, making the task much more challenging and requiring high-accuracy manipulation. Moreover, the movable mount needs to stay as close as possible to the original position to mimic a restricted working platform environment. The two fingers are mounted on a manipulator base that is allowed to move in the horizontal plane. The cost of this task is computed as:
    \begin{align}
        \mathcal{L} = \sum_{t = 1}^T &c_{mount} \|p^M_t - p^M_0\|^2 + c_{touch} (\|p^{left}_t - p^M_t\|^2 + \|p^{right}_t - p^{box}_t\|^2) \nonumber \\
        &+ c_p \|p^{box}_t - p_{hole}\|^2 + c_{rotation} \|\theta^M_t - \theta^{box}_t\|^2 \\
        &\text{with } c_{mount} = 15, c_{touch} = 1, c_p = 5, c_{rotation} = 50 \nonumber
    \end{align}
    where $p^M_t$ is the position of the movable mount at time $t$, $p^{left}_t$ and $p^{right}_t$ are the finger tip positions of left finger and right finger at time $t$, $p^{box}_t$ is the position of the small box at time $t$, $\theta^M_t$ and $\theta^{box}_t$ are the rotation angle of the mount and the box. The first term is used to penalize moving the mount too far away from the original place, the second term is designed to encourage the fingers to touch on the objects (but not indicate any specific position on the object), and the third term and the last term together is to measure how well the box is inserted into the mount.

    \item \textit{Free-form Gripper}: In this task, the algorithm needs to optimize the shape of two free-form gripper fingers in order to grasp a diamond-like object. The gripper fingers are cubes in the initial configuration but allowed for free-form deformation. We shake the object at the end of the episode to test the stability of the grasp. The control sequence in this task is pre-defined and fixed. The cost is computed by
    \begin{align}
        \mathcal{L} = \sum_{t = 1}^T h_t
    \end{align}
    where $h_t$ is the height of the target object.
\end{enumerate}

All the coefficients in the loss functions are selected to balance the importance and the value magnitude of different terms.

\subsection{More Experiment Results}
\begin{figure}[h!]
    \centering
    \includegraphics[width=\linewidth]{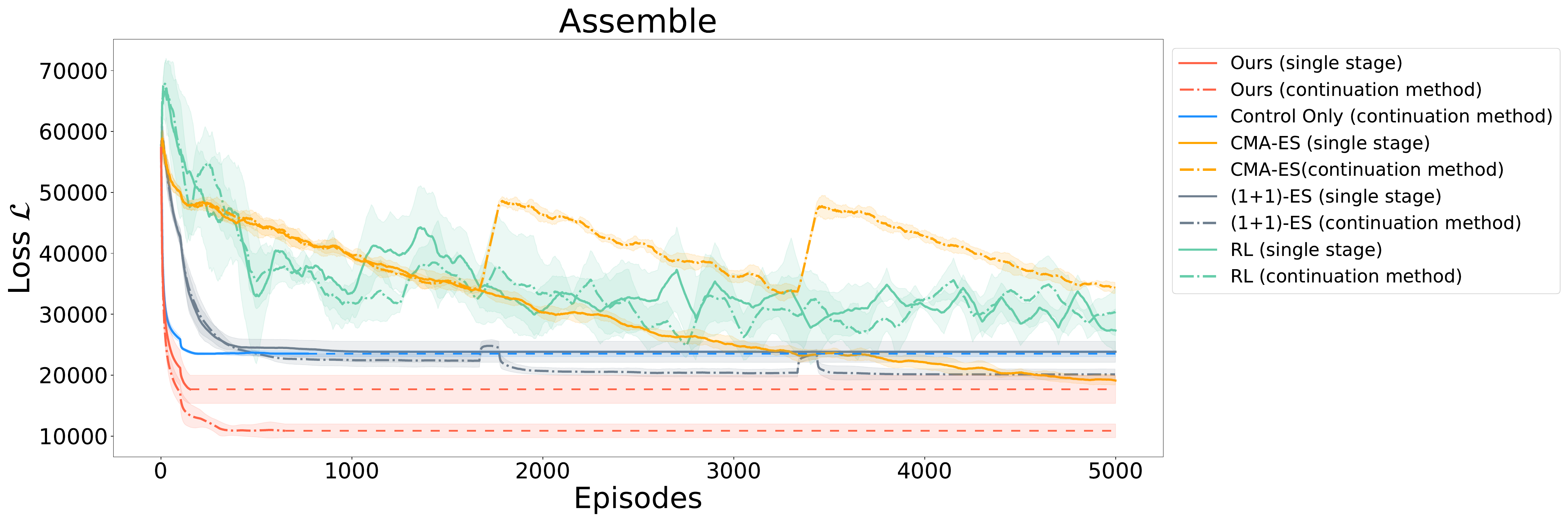}
    
    \caption{\textbf{Full plot of comparison curves for \textit{Assemble} task}. The curves are smoothed for better visualization and comparison.}
    \label{fig:full_comparison_assemble}
\end{figure}

We provide the full optimization loss curves for \textit{Assemble} task. In the \textit{Assemble} task, we found the continuation method~\cite{allgower2003continuation,geilinger2020add} can further reduce the task cost with our method. More specifically, we scale down the contact force to provide the optimization a smoother function objective space, and then scale the contact force up to the true value. we set a curriculum for the contact scale in $3$ stages. We set the contact scale to be $0.01$ in the first stage, $0.1$ in the second stage, and $1$ in the third stage. In our method, we start the next stage as long as the convergence is detected by L-BFGS-B for the previous stage. Scaling down the contact force initially allows the optimization to find a good initial solution for the final harder task. To make a fair comparison with baselines, we trained all the baselines with the continuation methods as well. We split the entire $5000$ training episodes into $3$ stages evenly for the baselines. As shown in Figure \ref{fig:full_comparison_assemble}, our method achieves significantly faster convergence and lower loss than all the baselines with or without the continuation method. While the continuation method speeds up the optimization and leads to lower converged loss using our method, it does not help improve the baselines' optimization performance. It even slows down the optimization in the case of CMA-ES. We hypothesize that it is because the baselines cannot solve the task (even the easiest \textit{Assemble} task with contact scale $0.01$) within a small number of iterations. Hence, the curriculum does not provide any benefit to the optimization.

\begin{figure}[h!]
    \centering
    \includegraphics[width=0.3\linewidth]{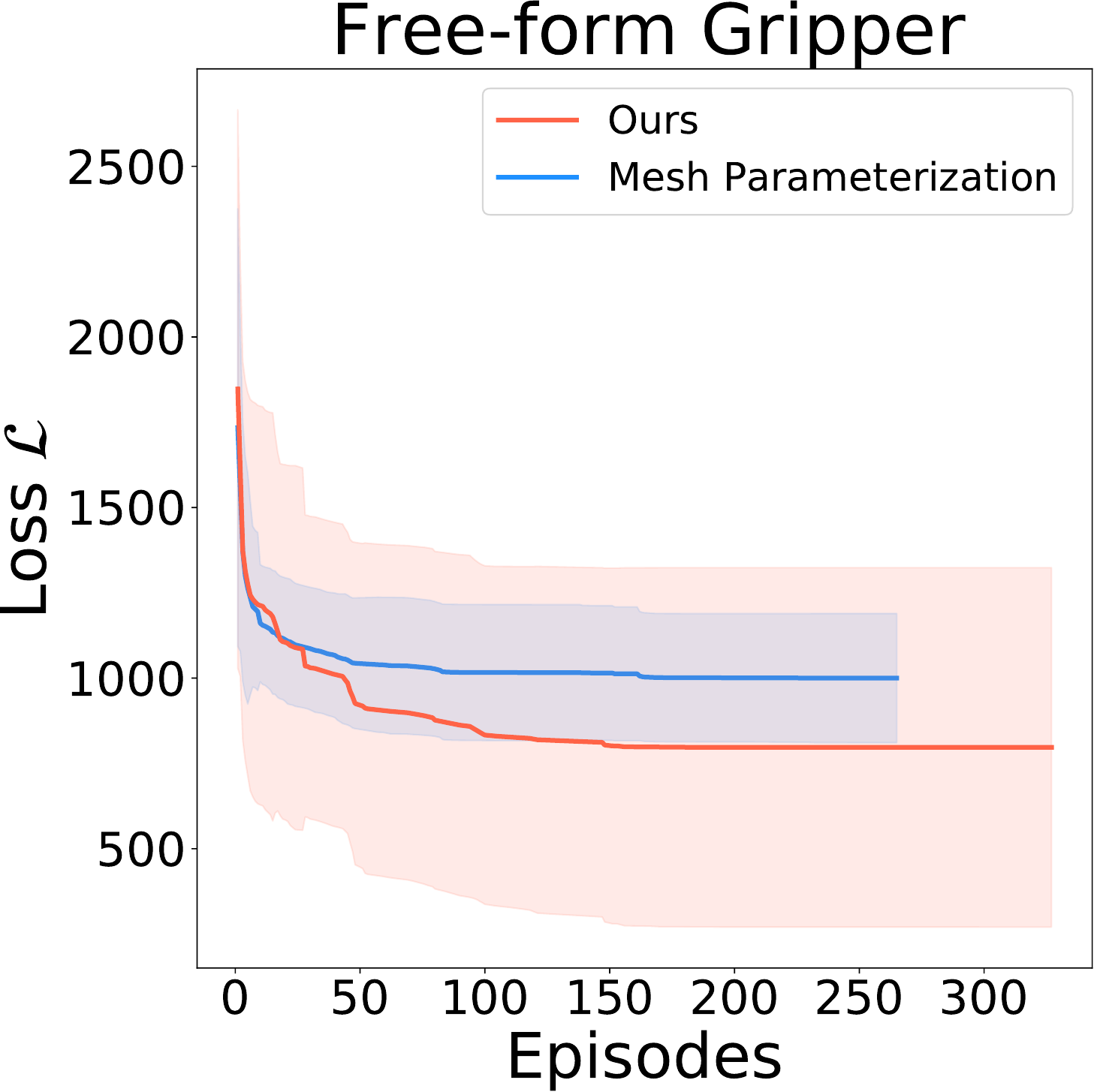}
    
    \caption{\textbf{Optimization curve comparison for \textit{Free-form Gripper} task}. The horizontal axis is the number of simulation episodes during optimization, and the vertical axis is the loss value. The experiment results are averaged from $30$ independent optimization runs with different initial guesses.}
    \label{fig:freeform_curve}
\end{figure}

We also show the loss curves of our deformation-based parameterization method and mesh-based parameterization method on the \textit{Free-form Gripper} task. We run each method for $30$ independent times with random initial morphology guesses and plot the average optimization loss curve in Figure \ref{fig:freeform_curve}.  Although our method fails in one optimization trial and the mesh parameterization succeeds in grasping the object in all $30$ optimization trials, we can see that our cage-based parameterization leads to better average performance. This is because our reduced morphology parameter space allows the optimization to explore more easily than the high DoF parameter space provided by the mesh parameterization method.

\end{document}